\documentclass[lettersize,journal]{IEEEtran}
\usepackage{amsmath,amsfonts}
\usepackage{algorithmic}
\usepackage{algorithm}
\usepackage{array}
\usepackage[caption=false,font=normalsize,labelfont=sf,textfont=sf]{subfig}
\usepackage{textcomp}
\usepackage{stfloats}
\usepackage{url}
\usepackage{verbatim}
\usepackage{graphicx}
\usepackage{cite}
\usepackage{enumitem}
\usepackage{orcidlink}
\usepackage{amssymb}
\usepackage{amsmath}

\usepackage{bbding}
\usepackage{pifont}

\usepackage{epsfig}
\usepackage{amsmath}

\usepackage{amssymb}
\usepackage{bbm}
\usepackage{booktabs, multirow}

\usepackage{xcolor}

\usepackage{paralist}
\graphicspath{{figures/}}

\usepackage{graphicx}
\usepackage{multirow}
\usepackage{multicol}
\usepackage{caption}
\usepackage{paralist}

\usepackage{algorithm}
\usepackage{algorithmic}

\newcommand{\lb}[1]{{\color{black}#1}}
\newcommand{\lbreb}[1]{{\color{black}#1}}

\newcommand{\lbnew}[1]{{\color{black}#1}}

\begin{document}

\title{Active View Selection for Scene-level Multi-view Crowd Counting and Localization with Limited Labeling Budget}

\author{Qi Zhang,~\IEEEmembership{Member,~IEEE,}
Bin Li, Antoni B. Chan,~\IEEEmembership{Senior Member,~IEEE,}
and Hui Huang*,~\IEEEmembership{Senior Member,~IEEE}

\thanks{Qi Zhang, Bin Li, and Hui Huang are with the College of Computer Science and Software Engineering, Shenzhen University, Shenzhen, China. Antoni B. Chan is with the Department of Computer Science, City University of Hong Kong, Hong Kong SAR, China.
E-mail: qi.zhang.opt@gmail.com, azswerfr@gmail.com, abchan@cityu.edu.hk, hhzhiyan@gmail.com}
\thanks{*Corresponding author: Hui Huang}
\thanks{Manuscript received xx, 2026; revised xx, 20xx}
}
 
\markboth{Journal of \LaTeX\ Class Files,~V ol.~xx, No.~xx, xx~2026}%
{Shell \MakeLowercase{\textit{et al.}}: A Sample Article Using IEEEtran.cls for IEEE Journals}

\IEEEpubid{0000--0000/00\$00.00~\copyright~2026 IEEE}

\maketitle


\begin{abstract}
Multi-view crowd counting and localization fuse the input multi-camera views for estimating the crowd number or locations on the ground. Existing methods mainly focus on accurately predicting on the crowd shown in the input views, which neglects the problem of choosing the `best' camera views to perceive all crowds well in the scene. Besides, existing view selection methods require massive labeled views and images, and lack the ability for cross-scene settings, reducing their application scenarios. Thus, in this paper, we study the view selection issue for better scene-level multi-view crowd counting and localization results with cross-scene ability and limited label demand, instead of input-view-level results. We first propose a baseline view selection method (IVS) that considers view and scene geometries in the view selection strategy and conducts the view selection, labeling, and downstream tasks independently. Based on the IVS baseline, we put forward an active view selection method (AVS) that jointly 
conducts
the view selection, labeling, and downstream tasks. In AVS, we actively select the labeled views and consider both the view/scene geometries and the predictions of the downstream task models in the view selection process under a limited labeling budget. Experiments on multi-view counting and localization tasks demonstrate the cross-scene and the limited label demand advantages of the proposed active view selection method (AVS), outperforming existing methods and with wider application scenarios.
\end{abstract}


\section{Introduction}
\label{sec:intro}

\begin{figure}[t]
\centering
\begin{center}
   \includegraphics[width=\linewidth]{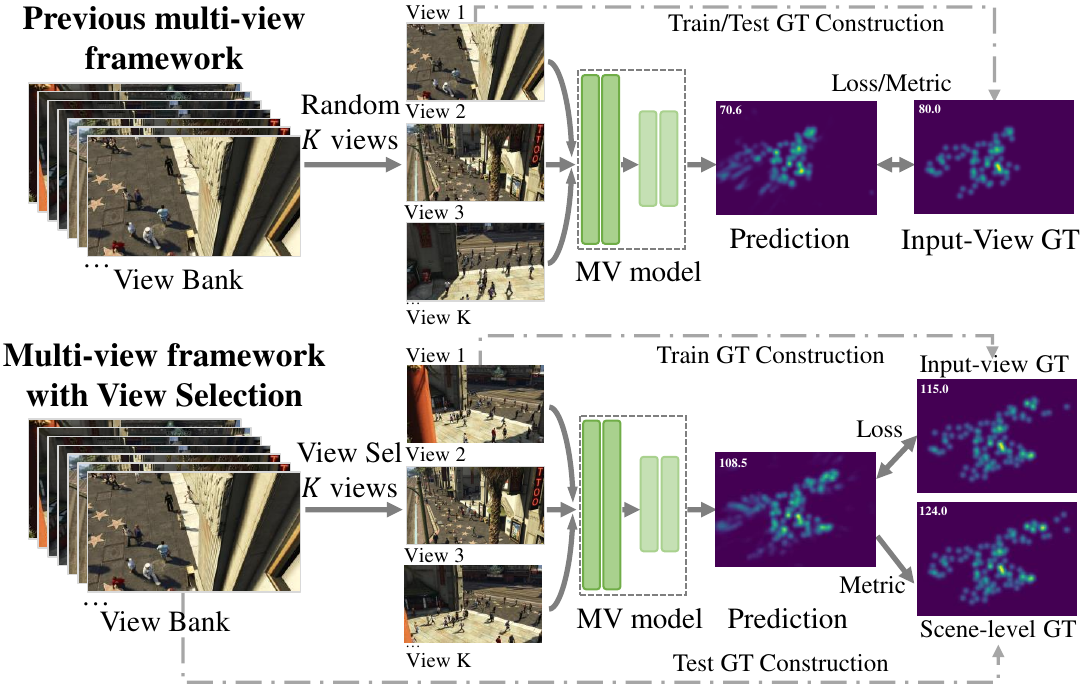}
\end{center}
   \caption{ 
   The comparison of the existing multi-view counting/localization frameworks and the scene-level multi-view framework with view selection.
   }
\vspace{-0.5cm}
\label{fig:general_idea}
\end{figure}







Multi-view crowd counting and localization leverage multiple camera views to predict the crowd counts or locations on the ground, alleviating the issue of severe occlusions in large and wide scenes. However, existing multi-view crowd counting and localization methods mainly focus on designing models for accurate estimation of the crowd covered by a randomly selected set of input views~\cite{zhang2021cross}, which may not contain or perceive all the crowds well in the scene, resulting in an incorrect prediction of the crowd in the scene. As in Fig.~\ref{fig:general_idea} top, these frameworks are trained and tested using the ground truth (GT) constructed from the randomly selected views, \textit{i.e.}, not tested on the whole scene.

Thus, for a complete multi-view vision system, we not only need to design better downstream task models (\textit{e.g.} counting, localization) but also select the best views for better scene-level downstream task performance. 
A simple two-stage solution is to conduct the view selection first, label the selected views, and then train the downstream multi-view models based on the labeled views, called independent view selection. 
As shown in Fig.~\ref{fig:general_idea} bottom, the view-selection-based multi-view framework is trained with GT constructed with selected views, and tested with the scene-level GT, where \textit{the GT constructed with selected views should be close to the scene-level GT}. 
\lbnew{
Recent path planning methods~\cite{xiong25drone,tang25drone,liang2025optimized} adopted the scene/view geometries in view selection for better 3D reconstruction.
}

\IEEEpubidadjcol

Unfortunately, few research works have studied the view selection issue in multi-view counting and localization for whole scene performance. In addition, the two-stage solution divides the view selection and downstream task model training into 2 separate stages, where the priors used for view selection may not be optimal settings for the downstream tasks. Furthermore, to reduce the annotation effort, sometimes only limited views are labeled, causing extra difficulties for downstream model learning. A recent method MVSelect~\cite{hou2024learning}~proposed a reinforcement learning (RL) framework for view selection and multi-view tasks. However, MVSelect requires GT annotations of \emph{all views}, making it impractical for selecting among large numbers of views to save annotation budget, and has a weak generalization ability due to the limitations of the proposed framework, \textit{making it not applicable to novel scenes}. 

\begin{figure*}[t]
\centering
\begin{center}
   \includegraphics[width=1\linewidth]{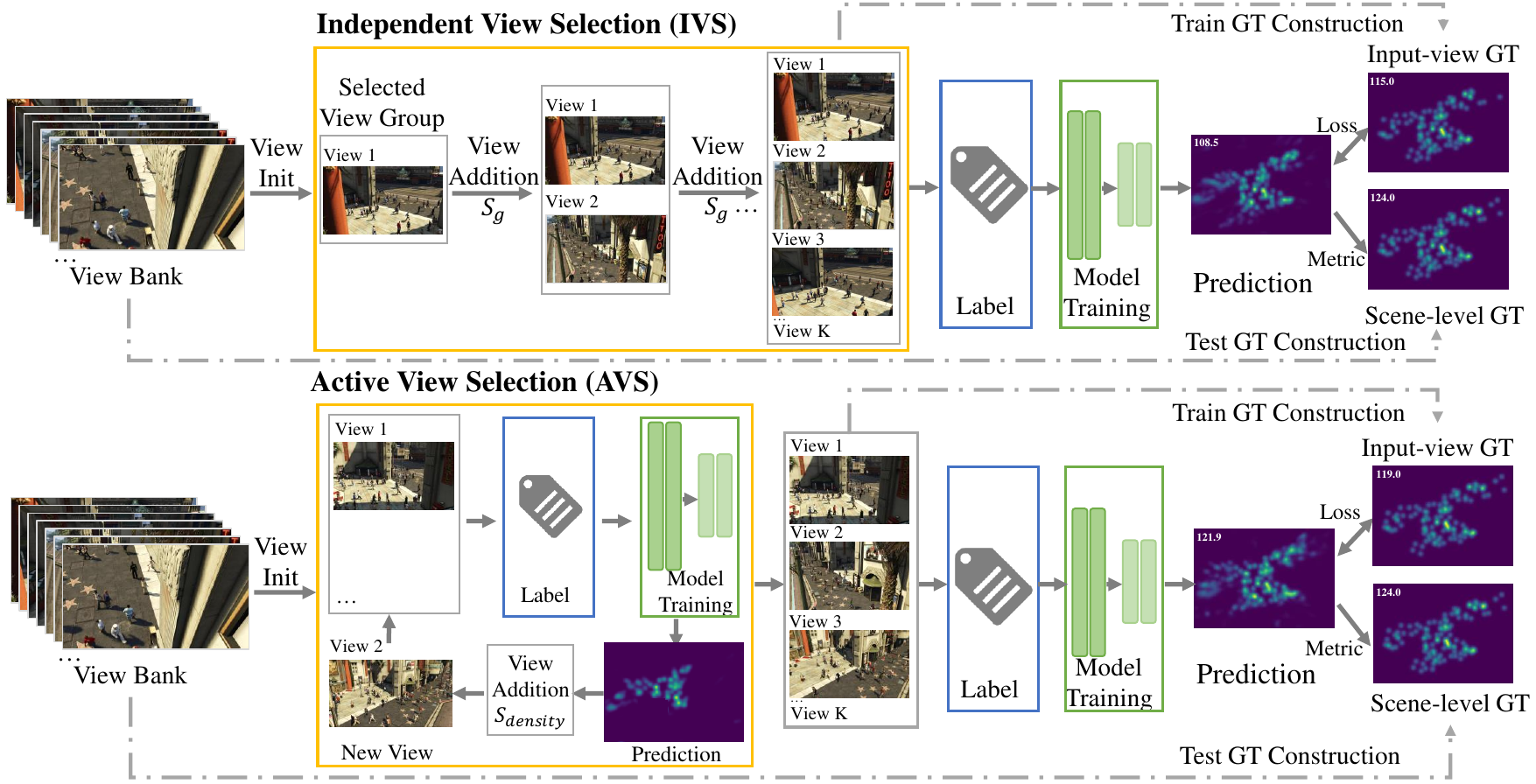}
\end{center}
\caption{
The pipeline of the proposed independent view selection baseline (IVS) and active view selection framework (AVS). IVS (top) separates the view selection (selecting views one by one with view selection score $S_g$ based on scene/view geometries) and downstream model training, while AVS (bottom) jointly conducts view selection and downstream model training: In the view selection process, the downstream model's prediction together and scene/view geometries are both used in view selection score $S_{density}$ to select new views, then the downstream model is trained with the updated 
view group, repeating the process until finishing selecting $K$ views, finally the downstream model is trained again with the selected and labeled views.}
\vspace{-0.4cm}
\label{fig:pipeline}
\end{figure*}

To address the mentioned issues, in this paper, we first propose an independent view selection \textbf{baseline} (IVS), based on which we further put forward an active view selection framework (AVS), requiring only limited labeled views and with cross-scene abilities.
IVS proposes a view selection score equation $S_g$ based on view and scene geometries, including 3 terms: \textit{the scene coverage, the average distance, and the view diversity}, working together to cover the crowd in the scene mostly and clearly.
The 3 terms are combined to select views that cover most scene areas to contain as many crowds as possible, have the lower average person-to-camera distance to ``see'' the crowd clearly, and have the higher view diversity instead of placing all cameras at the same locations. 
As illustrated in the top part of Fig.~\ref{fig:pipeline}, we first perform view selection to expand the selected view set from a single view to $K$ views based on the proposed view selection score equation $S_g$. The selected views are then annotated and used to train the multi-view task model. During training, GT is constructed from the selected input views, whereas during testing, the GT is derived from all available views.

Furthermore, in contrast to optimizing the view selection and the downstream model independently as in IVS, AVS jointly 
optimizes
the view selection and the downstream tasks by introducing the downstream task predictions in the view selection process. 
As shown in Fig.~\ref{fig:pipeline} bottom, the view selection score $S_{denisty}$ considers both the view/scene geometries (similar in IVS) and the prediction results of the downstream task models when expanding the selected view group 
from 1 to $K$ views, after which the downstream model is trained with the labeled selected views. 
Thus, the view selection, the data labeling, and the downstream model training are jointly conducted. 
Once $K$ views have been selected, the downstream model is retrained on the labeled views.
Additionally, to reduce annotation costs, we further introduce a novel pseudo-labeling strategy that is applied during both the view selection and downstream training stages.

The contributions of this paper are summarized as follows.
\begin{compactitem} 
    \item Few research works have studied the view selection problem for scene-level multi-view crowd counting and localization. We propose a novel independent view selection baseline IVS for scene-level downstream tasks, utilizing view and scene geometries in the view selection.
    \item Based on IVS, we propose an active framework AVS, where the view selection step and the downstream task models are jointly 
    optimized,     
    with better performance than IVS. 
    And we only require limited view labels from the selected views, via adopting pseudo labels on candidate views in the training. 
    \item  Experiments demonstrate that our method can apply to novel new scenes, with wider application scenarios, and outperform comparison methods on both multi-view counting and localization tasks.
\end{compactitem}





\section{Related Work}
\label{sec:relatedwork}

\begin{table*}[t]
\centering
\small
\caption{{Summary of main notations.}}
\begin{tabular}{cccc}
\hline
     Symbol    & Meaning      & Symbol   & Meaning \\
\hline

$F$              & The number of selected frames                                         	& $H_i$          & Visible scene region by view $i$                                  \\
$K$              & The number of selected views                                          	& $h$            & Scene height                                                      \\
$v_{max}$        & The view with the largest FOV                                         	& $w$            & Scene width                                                       \\
$V_{select}$     & Selected view group                                                   	& $D_p$          & Sum of inverse distance                       					 \\
$S$              & View selection score equation                                                & $S_{sc}$       & $S$ with scene coverage                   					 \\
$S_g$            & $S$ with view/scene geometries                           			 	& $S_{ad}$       & $S$ with average distance                     					 \\
$S_{mask}$       & Mask-indicated $S$                                   				 	& $S_{vd}$       & $S$ with view diversity                       					 \\
$S_{density}$    & Density-indicated $S$                              					 	& $B_{k}$        & Crowd density map mask                        					 \\
$N$              & Downstream task model                                                        & $M_k$          & Crowd density map                         					 \\
$\emptyset$      & No $N$                                                                       & $D^{den}_p$    & $D_p$ with $M_k$                          					 \\
$H_s$            & Scene region                                                          	& $V^k_{select}$ & $k$ selected views                                                \\
$H^k_v$          & Visible scene region by $V_{select}$                                  	& $V^g$          & View set of scene's $g$  										 \\
$G$              & Scene set                    										        & $E$            & The number of training epoch                                  \\
$V^g_{select}$   & Selected views of scene $g$  										        &                &  \\

\hline
\end{tabular}
\label{table:summary_notation}
\end{table*}


\textbf{Multi-view crowd counting.}
Compared to single-image counting~\cite{liang2022end,han2023steerer,zhao2020active,savner2023crowd,zhang2024boosting,lin2025point,zhou2025crowd,shang20262d}, multi-view crowd counting is proposed to handle scenes with large areas, severe occlusions, or irregular shapes by fusing multiple synchronized and calibrated camera views.
Traditional methods~\cite{Ryan2014Scene,Tang2014Cross} employed foreground extraction techniques and hand-crafted features, with limited performance and generalization abilities.
Recently, deep learning methods~\cite{zhang2019wide,zheng2021learning,zhai2022co,WSFCMVCC123} 
have been introduced, trained, and tested with ground-plane density maps based on the crowds appearing in the input camera views.
CVCS~\cite{zhang2021cross}  proposed a camera selection model for cross-view cross-scene multi-view counting with a large synthetic cross-scene multi-view dataset. 
~\cite{mo2024countformer} put forward a transformer model with attention-mechanism-based 2D-3D feature lifting. 
Overall, existing multi-view counting methods focus on the accurate crowd number estimation of the people contained in a randomly selected set of input camera views. 
\textit{Current SOTAs have not yet explored the problem of selecting the best views for scene-level multi-view counting}.
\lbreb{
Moreover, existing pretrained single-image models can serve as backbones for multi-view models, thereby improving training speed and effectiveness. And it can directly help us obtain crowd information to enhance perception, such as the initialization process of the proposed methods.
}

\textbf{Multi-view crowd localization.}
Multi-view crowd localization estimates the crowd locations on the ground in the scene.
Early methods' performance is limited~\cite{chavdarova2017deep,baque2017deep} due to no view feature alignment.
Recent methods~\cite{song2021stacked,hou2021multiview,qiu20223d,liu2024unsupervised,zhang2024multi,zhang2024mahalanobis,aung2025multi,ma2025dchm,alturki2025enhanced} put forward end-to-end frameworks with better performance.
MVDet~\cite{hou2020multiview} used feature perspective transformations to fuse multi-views. \cite{aung2025multi} extends multi-view crowd localization to pedestrian occupancy prediction, achieving SOTA performance on the proposed benchmark.
CaMuViD~\cite{etefaghi2025camuvid} facilitates flexible transformation and improves feature fusing across views, removing the need for BEV representation and achieving better detection accuracy. 
Similarly, \textit{most SOTA multi-view crowd localization methods also focus on estimating crowds `seen' in the input camera views, not targeting all crowds in the scene}. 
MVSelect~\cite{hou2024learning} is the most related to our paper, and it proposed a reinforcement learning (RL) framework for view selection and downstream tasks. However, MVSelect requires annotations of all views to train the model, and has a weak generalization ability to apply to novel scenes. 
\textit{In contrast, our method only needs to label the selected views, and with the aid of the proposed pseudo labels, it can be well applied to novel news scenes in the test stage} 
(see experiments).

\textbf{View selection for other multi-view tasks.}
View selection is also vital in many other multi-view vision tasks~\cite{luo2026vtr,majumder2025viewpoint,di2025learning,li2025nextbestpath,kiciroglu2020activemocap,zheng2024gps,sun2021learning,ruan2023towards,Du_2023_ICCV}, such as path planning, or multi-view object classification.
~\cite{liu2022learning} measured the reconstructability in a learning way and designed an interactive path planning framework for view selection.
MVTN~\cite{hamdi2021mvtn} directly regresses optimal viewpoints for 3D shape recognition with an MLP.
\lb{
\cite{Du_2023_ICCV} proposed a reinforcement learning-based framework for multi-view active fine-grained visual recognition. 
\cite{xiao2024nerf} proposed a unified framework for view selection methods and devised a thorough benchmark to assess its impact on neural rendering.
}
\lbnew{
By considering the reconstruction errors’ spatial distribution of the 3D predictions, a reconstruction error-guided view selection method~\cite{zhang2025view} has been proposed and achieves better reconstruction performance.
\cite{xiong25drone} designs a comprehensive texture quality assessment system to guide view planning, and proposes an innovative aerial path planning framework designed to co-capture images for reconstructing both structured geometry and high-fidelity textures.
Furthermore, \cite{tang25drone} introduces a novel changeability heuristic to evaluate the likelihood of scene changes, driving the planning of two flight paths to select better viewpoints for 3D reconstruction.}
\textit{It is a trend in other multi-view tasks to jointly conduct the view selection and the downstream task. However, there is little research on view selection for scene-level multi-view crowd counting and localization tasks}, which is a relatively unexplored area.

\section{Active View Selection Framework}
\label{sec:method}

\lbreb{
We first propose a novel independent view selection \textbf{baseline} (IVS) adopting a two-stage process for scene-level downstream tasks. 
Next, based on IVS, 
we propose the active view selection framework (AVS) for jointly optimizing view selection and downstream task model training.
Pseudo labels are adopted to enhance the model's cross-scene generalization abilities. 
For both IVS and AVS, we assume an annotation budget of $F$ frames per view and $K$ views of each scene, or a total $FK$ images per scene. We provide a summary table of the main notations
shown in Table \ref{table:summary_notation} for symbol querying.

}

\subsection{Independent View Selection Baseline (IVS)}
In IVS, we first initialize the selected frames and the first view, then start to add new views one by one with the proposed view selection score equation based on scene/view geometries, expanding the selected view number from 1 to $K$.

\subsubsection{Initialization}

Initialization has two stages: First, selecting the $F$ frames to be processed and selecting the first view.
For frame selection, we first find the view $v_{max}$ with the largest field-of-view (FOV) area on the ground. Then, we select the first frame as the one with the largest predicted crowd count in view $v_{max}$ using DM-Count~\cite{wang2020distribution},
\lbreb{
a pre-trained single-image counting model that can help effectively perceive crowd information in a label-efficient manner.
}
Next, given the diversity in the limited labels, we select the rest frames with the lowest cosine similarity between the selected frames and candidate frames of view $v_{max}$. This process is repeated until $F$ frames are selected. 
For view initialization, the view with the largest crowd count sum across all selected $F$ frames is selected as the first view.

\begin{algorithm}[t]
\caption{Independent View Selection Framework}
\label{alg:independent_view_selection}
\begin{algorithmic}[1]
\small
\STATE \textbf{Input}: each scene's total views $V^g\!\in\!\{v_{1}^{g},...,v_{n}^{g}\}$, all the scenes $G\!=\!\{g\}$, max selected view number $K$, view selection score equation $S=S_g$, task model $N$.
\STATE initialize frames and the selected view group $\{V_{select}^{g}\}$.
\FOR{$g \in G$}
    \FOR{$k \in \left\{2,\ldots,K\right\}$}
        \STATE view\_addition($V^g, V_{select}^{g}$, $S$, N);
    \ENDFOR
\ENDFOR
\STATE label all the selected views $\{V_{select}^{g}\}$;
\STATE model\_training($N,\{V_{select}^{g}$\}).
\end{algorithmic}
\end{algorithm}

\begin{center}
    
\begin{algorithm}[t]
\caption{View Addition}
\label{alg:add_view}
\begin{algorithmic}[1]
\small
\STATE \textbf{Input}: all views $V^g \in \{v_{1}^{g},...,v_{n}^{g}\}$ of scene $g$, selected view group $V_{select}^{g}$ of scene $g$, view selection score equation $S \in \{S_{g},S_{mask},S_{density}\}$, and task model $N$ ($=\emptyset$ if $S=S_g$).
\STATE ${s}\_v_{select}=-\inf$;
\FOR{$v \in V^g\setminus V_{select}^g $}
    \STATE $s\_{v}=S(\{V_{select}^{g}, v\}, N) $;
    \IF{$s\_{v} > {s}\_v_{select}$}
        \STATE $v_{select}=v$;
        \STATE ${s}\_v_{select}={s}\_v;$
    \ENDIF
\ENDFOR
\STATE $V_{select}^g=\{V_{select}^g,v_{select}\}$.

\end{algorithmic}
\end{algorithm}
\end{center}

\subsubsection{View Addition: $S_g$}
\lbreb{
With the selected frames generated by frame initialization and the selected view group $V_{select}$ including the first view from view initialization above, a view selection score equation $S_g$ is proposed for view addition.
}
For each iteration, the $S_g$ score of each candidate view together with the current $V_{select}$ is calculated, and then we select the new view with the largest score. The process is repeated until the specified $K$ views are selected, as shown in Algorithm \ref{alg:add_view}.
$S_g$ consists of 3 terms: Scene Coverage, Average Distance, and View Diversity, which are as follows.


\begin{figure}[t]
\centering
   \includegraphics[width=\linewidth]{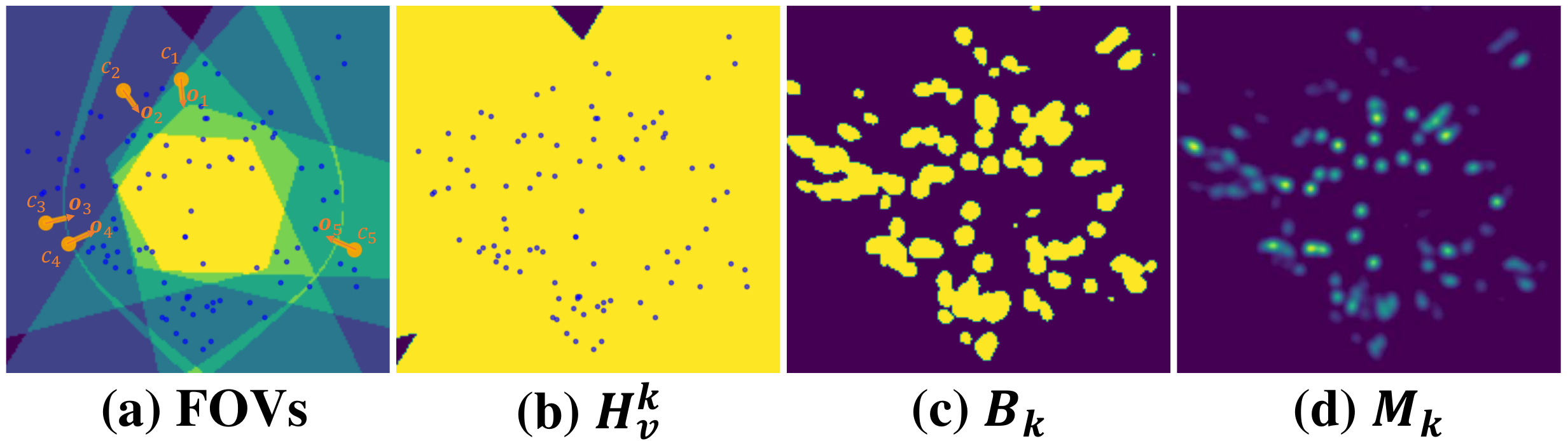}
   \caption{ 
   (a) The combined FOVs of selected views; (b) The FOV mask; (c) The crowd mask; (d) The crowd density. The dots in (a) and (b) are ground-truth crowd locations.
   }
\vspace{-0.6cm}
\label{fig:score_term}
\end{figure}


\textbf{Scene coverage} score term $S_{sc}$ indicates whether the selected views can cover all crowds in the scene as much as possible. 
We first use the scene ground plane map as the scene region $H_s$, whose area size is $Area(H_s)=hw$, and $h$ and $w$ are the height and width of the map. 
Then, we calculate the visible scene areas covered by the selected views as $H^k_v=\{H_1\cup H_2  ...H_k\}$, which is the combined FOV region of $k$ views (see Fig.~\ref{fig:score_term} (a) and (b)), and $H_i$ is view $i$'s FOV covering region. 
Thus, the area ratio of the visible region $H^k_v$ and the scene region $H_s$ is defined as $S_{sc}$:
\begin{align}
    S_{sc} = \sum H^k_v/ Area(H_s) =   \sum \{H_1\cup H_2  ...H_k\} /(hw), \label{S_sc}
\end{align} 
\lb{
where a larger $S_{sc}$ indicates a higher probability of covering all crowds by the selected views.
}

\textbf{Average distance} score term $S_{ad}$ considers the average person-to-camera distance in the view selection, indicating whether the selected views can `see'  the crowd clearly. 
Specifically, for each location $p$ in the visible region $H^k_v$, the person's inverse distance to the currently selected $k$ cameras is calculated as  $D_p = \textstyle \sum^k_{i=1} {1/{\left\lVert p-c_i \right\rVert}}$, where $c_i$ is the $i$-th camera's location on the ground (see Fig.~\ref{fig:score_term} (a)) and $p$ is in $H_i$, where higher values indicates shorter distance to the selected cameras. 
Combining all crowds, $S_{ad}$ is:
\begin{align}
    S_{ad} &= \frac{\sum_{p \in H^k_v} D_p}{\sum H^k_v}. \label{S_ad}
\end{align}
As the crowd locations are not known, all locations in the approximate visible region $H^k_v$ are used in the calculation. Similarly, a higher $S_{ad}$ forces the selected cameras to be close to the crowds and captures the crowds more clearly.

\textbf{View diversity} score term $S_{vd}$ avoids the setting that all cameras are located at the same place and point out.
Because we require multi-cameras to be placed at different corners facing each other to make full use of their occlusion handling potential~\cite{zhang2019wide}. Thus, we adopt a similarity measure~\cite{zhou2020offsite} to calculate $S_{vd}$, which is defined as follows:
\begin{align}
    S_{vd}  = \exp{(-\lambda \sum^k_{i=1} \sum ^k_{j=i+1} \frac{\textbf{o}_i \cdot \textbf{o}_j}{\lVert c_i-c_j \lVert+\epsilon})}, \label{S_vd}
\end{align}
where $\textbf{o}_i$ and $\textbf{o}_j$ are the camera optical axis directions (see  Fig.~\ref{fig:score_term} (a)), $c_i$ and $c_j$ are camera locations, $\lambda$ is a hypeparameter, and $\epsilon$ is a small value to avoid zero denominator. When the selected cameras have larger view direction and location differences, i.e., view diversity, $S_{vd}$ is higher.

By combining the 3 terms, we obtain the independent view selection score equation $S_{g}$ by only considering the view and scene geometries.
\begin{align}
    S_{g}  &= S_{sc}*S_{ad}*S_{vd} \\
           &= \frac{\sum_{p \in H^k_v} D_p}{ Area(H_s)} \exp{(-\lambda \sum^k_{i=1} \sum ^k_{j=i+1} \frac{\textbf{o}_i \cdot \textbf{o}_j}{\lVert c_i-c_j \lVert+\epsilon})}. \label{Sg}
\end{align}

Once the required frames and views are selected, they are annotated, and then the downstream task model is trained on the labeled data (see Fig.~\ref{fig:pipeline} top right). 
The main \textbf{weakness} of the independent view selection baseline (IVS) is that the view selection and downstream model training are separated, which does not ensure an optimal result for scene-level tasks. For example, the selected views are not necessarily suitable for downstream model training due to multi-view counting and localization models being sensitive to view angles, heights, or other properties. 

\subsection{Active View Selection (AVS)}

\begin{algorithm}[t]
\caption{Active View Selection Framework}
\label{alg:joint_view_selection}
\begin{algorithmic}[1]
\small
\STATE \textbf{Input}: each scene's total views $V^g \in \{v_{1}^{g},...,v_{n}^{g}\}$, all the scenes $G\!=\!\{g\}$, max selected view number $K$, training epochs $E$, threshold $\tau$ to add view, view selection score equation $S \in \{S_{mask},S_{density}\}$, and task model $N$. 
\STATE initialize frames and the selected view group $\{V_{select}^{g}\}$.
\STATE label all the selected views $\{V_{select}^{g}\}$.
\FOR{$e \in \left\{1,\ldots,E\right\}$}
    \STATE $metric$ = model\_training($N$,$\{V_{select}^{g}\}$);
    \IF{$ metric > \tau $ \AND {$ len(V_{select}^{g})<K $}} 
        \FOR{$g \in G$}
            \STATE view\_addition($V^g, V_{select}^{g},S, N$);
        \ENDFOR
        \STATE label all the added new views;
    \ENDIF
\ENDFOR
\end{algorithmic}
\vspace{-0.1cm}
\end{algorithm}

To address the weakness of the IVS baseline--optimizing the view selection and downstream model separately, we propose the AVS framework that jointly 
optimizes
the view selection and downstream task models as in Fig.~\ref{fig:pipeline} bottom. In the view selection process, the intermediate model's prediction together with the scene/view geometries is adopted in the view selection score $S_{density}$, and selected new views are labeled and used to train the downstream model, which is repeated until the desired view number is reached. Finally, the downstream model is trained again with the selected and labeled views.
The complete algorithm procedure is presented in Algorithm \ref{alg:joint_view_selection}. The initialization process is the same as IVS.
We update the view selection score equation in IVS by introducing the downstream task predictions, denoted as $S_{mask}$ and $S_{density}$, with details as follows.

\textbf{Mask-indicated view selection $S_{mask}$.}
The visible region in (\ref{Sg}) is defined by the combination of the FOVs of the selected cameras, which neglects the actual crowd regions in the scene. Therefore, we rely on the prediction density maps $M_k$ from the downstream model $N$ of the selected $k$ views $\{v_1, v_2, ..., v_k\}$ to more accurately indicate the appearing crowds' regions in the scenes: $M_k = N(v_1, v_2, ..., v_k)$. Specifically, we binarize $M_k$ with a threshold $\sigma$ to obtain a crowd density map mask $B_k$ (see Fig.~\ref{fig:score_term} (c)), which is used to replace $H^k_v$ in (\ref{S_sc}), (\ref{S_ad}) and (\ref{Sg}). Thus, we obtain
\begin{align}
    S_{mask} = \frac{\sum_{p \in B_k} D_p}{  Area(H_s)} \exp{(-\lambda \sum^k_{i=1} \sum ^k_{j=i+1} \frac{\textbf{o}_i \cdot \textbf{o}_j}{\lVert c_i-c_j \lVert+\epsilon})}, \label{S_mask} 
\end{align}
where $B_k$ indicates \textit{the crowd location information, which is utilized in the view selection process}. Note that the downstream counting or localization model $N$ is involved in the view selection score term $S_{mask}$, while the newly selected views during the view selection process could be fed into and train the downstream model. Therefore, the view selection and downstream task model are interacting in these two steps and thus influence each other.

\textbf{Density-indicated view selection $S_{density}$.} 
The mask-indicated view selection uses the binarized density map $B_k$ to indicate the crowd-visible regions, which neglects the crowd density's influence on the view selection score term. In other words, the crowded areas with higher densities should have higher weights in the view selection process. Therefore, we propose the density-indicated view selection score $S_{density}$ by introducing the density prediction $M_k$ (see Fig.~\ref{fig:score_term} (d)) of the downstream task model into $D_p$ in  (\ref{S_ad}), which is rewritten as $D^{den}_p = \textstyle \sum^k_i \frac{M_k(p)}{{\left\lVert p-c_i \right\rVert}}$,
where $M_k(p)$ indicates the density value of point $p$. Thus, \textit{by updating ${S_{ad}}$ with $D^{den}_p$, and replacing $H^k_v$ with $B_k$ in $S_{sc}$ and $S_{ad}$}, the view selection score term in  (\ref{S_mask}), we obtain:
\begin{align}
    S_{density}\! = \!\frac{\sum_{p \in B_k} \! D^{den}_p}{ \!Area(H_s)} \exp{(-\!\lambda \! \sum^k_{i=1} \!\sum ^k_{j=i+1} \! \frac{\textbf{o}_i \cdot \textbf{o}_j}{\lVert c_i-c_j \lVert+\epsilon})}. \label{S_density}
\end{align}
The view selection score term $S_{density}$ considers both \textit{the view/scene geometries, and the crowds' density level and location information}, where the view selection and downstream model training are conducted jointly.



\begin{figure}[t]
\centering
   \includegraphics[width=1\linewidth]{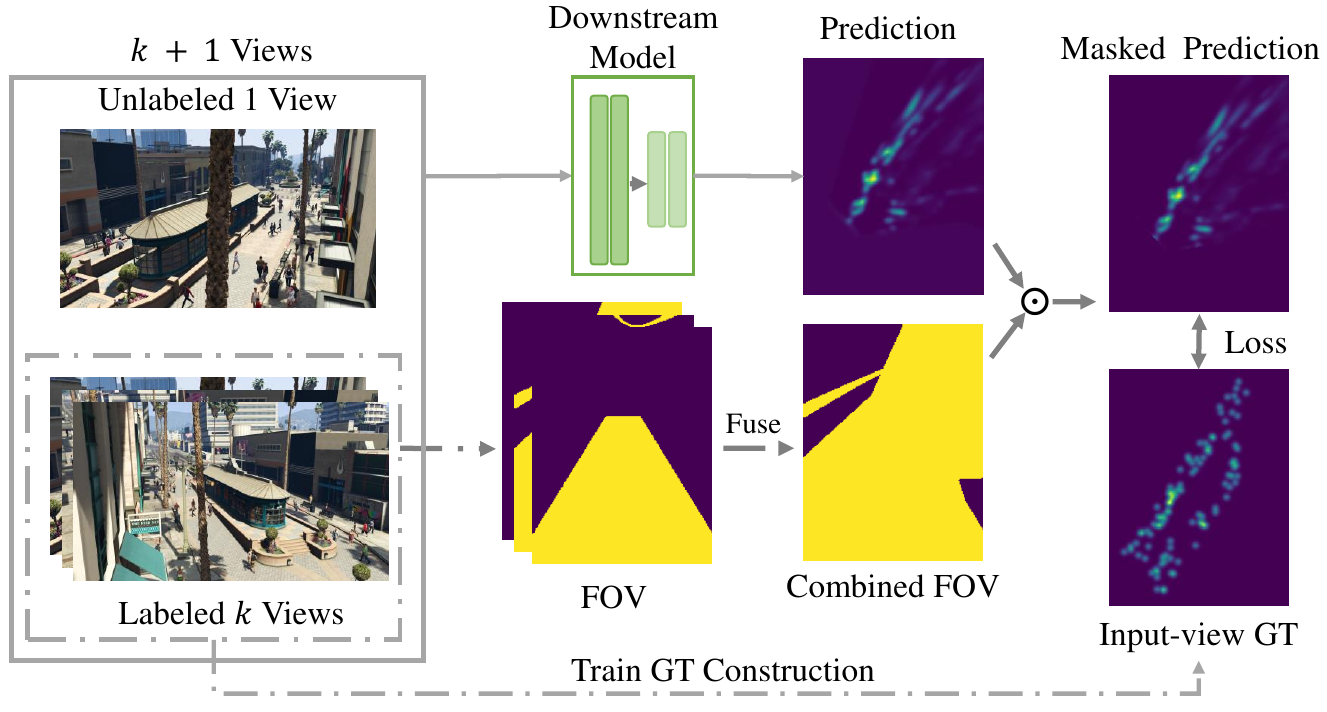}
   \caption{The pseudo label during view selection. $\odot$ denotes the element-wise multiplication of matrices.} 
\label{fig:pseudo_pipeline_during_view_selection}
\vspace{-0.4cm}
\end{figure}

\subsection{Pseudo Labels and Training}
{
To enhance the model's generalization ability, we utilize novel pseudo labels to train the downstream model better.
During view selection, the currently selected views $V^k_{select}$ and a random unselected view are combined as pseudo inputs to train the model, whose GT is ground-plane density maps of crowds covered by $V^k_{select}$.
Specifically, as in Fig.~\ref{fig:pseudo_pipeline_during_view_selection}, for the pseudo inputs in the view selection, labeled $k$ views and 1 random view from the remaining unlabeled views will be together as the model's inputs (total $k+1$ views) to obtain the predicted ground-plane density map, and the predicted density map is further masked by the combined FOV mask of the labeled $k$ views.
The corresponding GT ground-plane density map is constructed from the labeled $k$ views, which is accurate for supervising the prediction masked by the combined FOV of the selected $k$ views.
Thus, additional unlabeled views can be incorporated into downstream model training, thereby enhancing its generalization to new views. 

Besides, after the view selection, the selected views $V^K_{select}$ (note that $K$ is the final number of the selected views) of the $F$ selected frames are used for downstream model training. In addition to that, we also add pseudo inputs in training, which is a mix of $1$ selected view and $K\!-\!1$ unselected random views, whose pseudo-GT is the $K$ selected views' GT ground-plane density maps masked by the intersection of $H^K_v$ and the pseudo input views' combined FOV mask, and the prediction is also masked by the intersection FOV.
Specifically,  as shown in Fig.~\ref{fig:pseudo_pipeline_after_view_selection}, for the pseudo inputs after view selection, 1 selected view (random chosen from $V^K_{select}$) and $K\!-\!1$ random views from the remaining unlabeled views are combined and regarded as the model's inputs to predict the ground-plane density map. The GT ground-plane density map is constructed from the labeled $K$ views. 
Since the pseudo inputs and GT are from different views, we can only supervise the common regions covered by the pseudo input views and the $K$-labeled views constructing the GT. Thus, we add a FOV intersection mask on both the prediction and GT density map in the loss calculation, where the intersection mask is the common region of the combined FOVs of the pseudo-input views and the $K$-labeled views.
Note that both pseudo labels, which are used during the view selection stage or applied after view selection, are generated based on the selected $F$ multi-frames and the labeled views.


\begin{figure}[t]
\centering
   \includegraphics[width=1\linewidth]{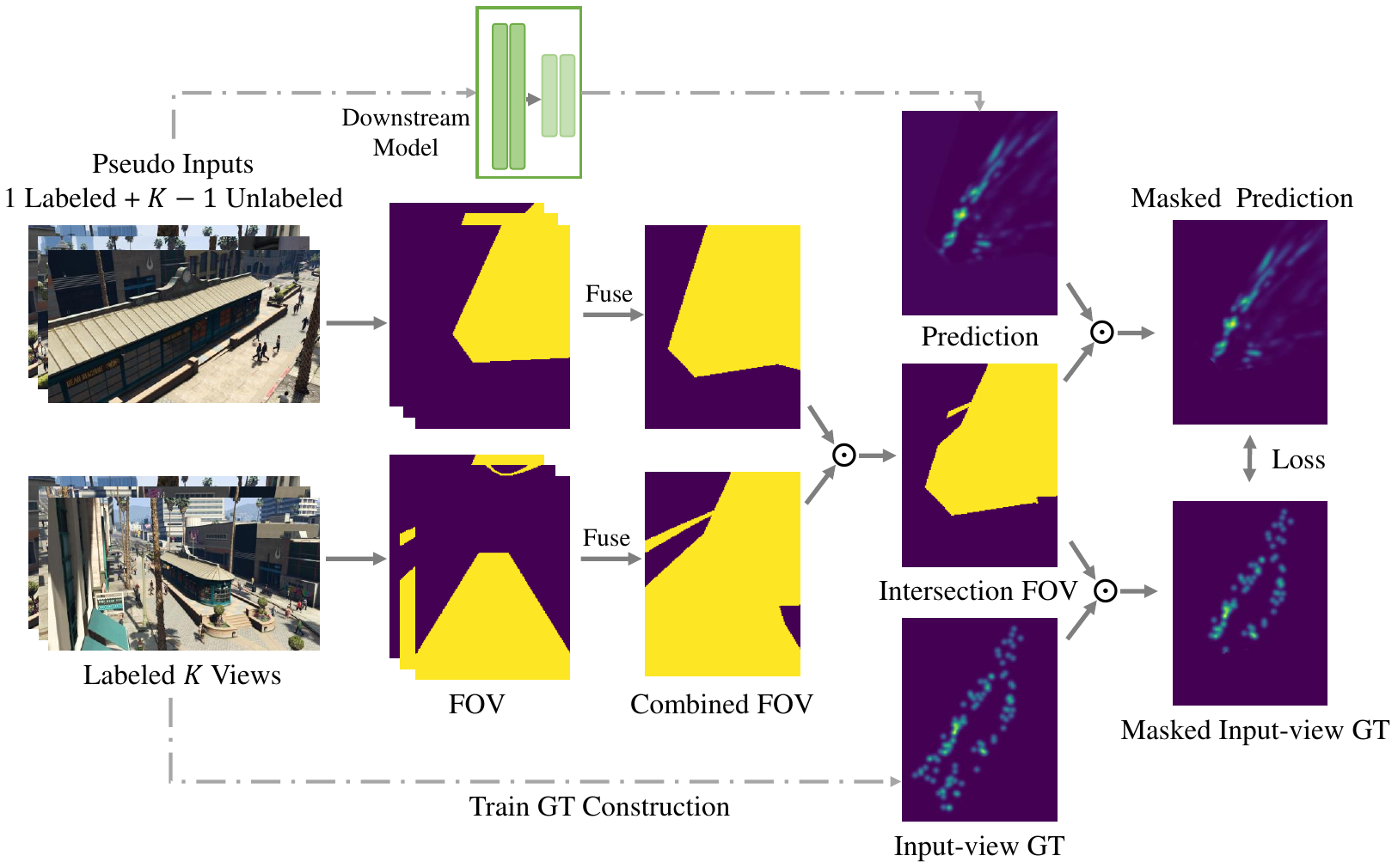}
   \caption{The pseudo label in the final downstream model training after view selection. 
   } 

\label{fig:pseudo_pipeline_after_view_selection}
\vspace{-0.4cm}
\end{figure}

\begin{table} [t]
\centering
\caption{The label cost of selected view number $K$ and selected $F$ on different datasets.} 
\begin{tabular} {lcccccc} 
\hline
Dataset & View & Scene & Original Training Frame & $K$  & $F$ \\ 
\hline
CityStreet  & 3                             & 1                    & 300                         & 2                                    & 60                          \\ 
CVCS  & 60-120                             & 31                    & 100                         & 5                                    & 20                          \\ 
Wildtrack  & 7                             & 1                    & 360                         & 3                                    & 360                           \\ 
MultiviewX & 6                             & 1                       & 360                         & 3                                  & 360                           \\ 
\hline
\end{tabular} 
\label{tab:cp3_dataset_cost} 
\end{table}

By using pseudo labels, a large number of unlabeled views are incorporated into model training, significantly enhancing the model’s generalization. \textit{Both IVS and AVS adopted pseudo labels in the model training.} 
For IVS, because it does not involve joint training with downstream tasks during view selection, the pseudo-labels are used only in the final training of the downstream model after view selection. Additionally, for both IVS and AVS, the training set samples are doubled during the final model training stage after view selection. Specifically, compared with the original training method, which doesn't use the pseudo inputs, the pseudo inputs are used as additional samples, \textit{i.e.} in the ratio of 1:1 for the $K$-labeled view inputs and the pseudo-label view inputs.

\section{Experiments and Results}
\label{sec:experiments}

\begin{table}[t]
\centering
\small

\centering
\small
\caption{Comparison of the multi-view counting results on the CVCS dataset. 
}
\begin{tabular}{l@{\hspace{0.12cm}}|c@{\hspace{0.12cm}}c@{\hspace{0.12cm}}c@{\hspace{0.1cm}}|c@{\hspace{0.1cm}}}
\hline
    Method                  & MAE $\downarrow$    & MSE  $\downarrow$      & NAE  $\downarrow$  & CoverRate $\uparrow$  \\
\hline

MVMS~\cite{zhang2019wide} (Random)                & 37.22 &	44.74 &0.276 &		0.872 \\
CVCS~\cite{zhang2021cross} (Random)               & 30.97 &	36.47 &0.228 &		0.901 \\
    Uniform
    & 21.76 & 25.75 & 0.163 &0.945 \\
    Random                                                  & 36.59 & 42.06 & 0.271 &0.885 \\
    Random (Pseudo)                                         & 28.22 & 33.73 & 0.208 &0.885 \\  
   MVSelect~\cite{hou2024learning}                          & 13.58 & 17.94 & 0.099 &0.899 \\
\hline
    IVS\_$S_g$ (Ours)       &  14.81     &   18.47    & 0.110                      &0.954          \\
    AVS\_$S_{mask}$ (Ours)    & 12.53 & 15.33 & 0.093                              &0.955          \\
    AVS\_$S_{density}$ (Ours) & \textbf{10.99}  & \textbf{13.57} & \textbf{0.083}  &\textbf{0.960} \\
\hline
\end{tabular}
\vspace{-0.4cm}
\label{table:CVCS_results}
\end{table}


\subsection{Experiment Settings}
\textbf{Experiment design.}
In the training, IVS conducts the view selection, labeling, and downstream model training independently, while AVS conducts them jointly until the view number reaches $K$ and then trains the downstream task model on the labeled $K$ views. In the testing, no model training is needed for IVS or AVS, where the same view selection process is conducted with all testing frames, and the downstream model prediction is directly used in the view selection score.

\textbf{Dataset.}
We use publicly available datasets
and comply with the original licenses and terms of use.
The multi-view crowd counting task is conducted on CVCS~\cite{zhang2021cross} and CityStreet~\cite{zhang2019wide} datasets. The multi-view crowd localization task is conducted on Wildtrack~\cite{chavdarova2018wildtrack} and MultiviewX~\cite{hou2020multiview} datasets.
\begin{compactitem}
    \item 
    \textbf{CVCS} is a multi-scene dataset which includes 23 scenes for training and 8 scenes for testing, with 60-120 camera views in each scene with calibrations, which is challenging and suitable to \textit{validate the cross-scene generalization of the proposed frameworks}. 
    \item 
    \textbf{CityStreet} is a single-scene real-world dataset and contains 3 camera views and 300 frames for training and 200 for testing. 
    \item 
    \textbf{Wildtrack} and \textbf{MultiviewX} are two single-scene datasets, where the same settings are used as in MVSelect~\cite{hou2024learning}: 360 frames are all used for model training and 40 frames are for testing, without frame selection. 
\end{compactitem}
The label costs of each dataset are shown in Table~\ref{tab:cp3_dataset_cost}.

\begin{figure*}[t]
\centering
   \includegraphics[width=\linewidth]{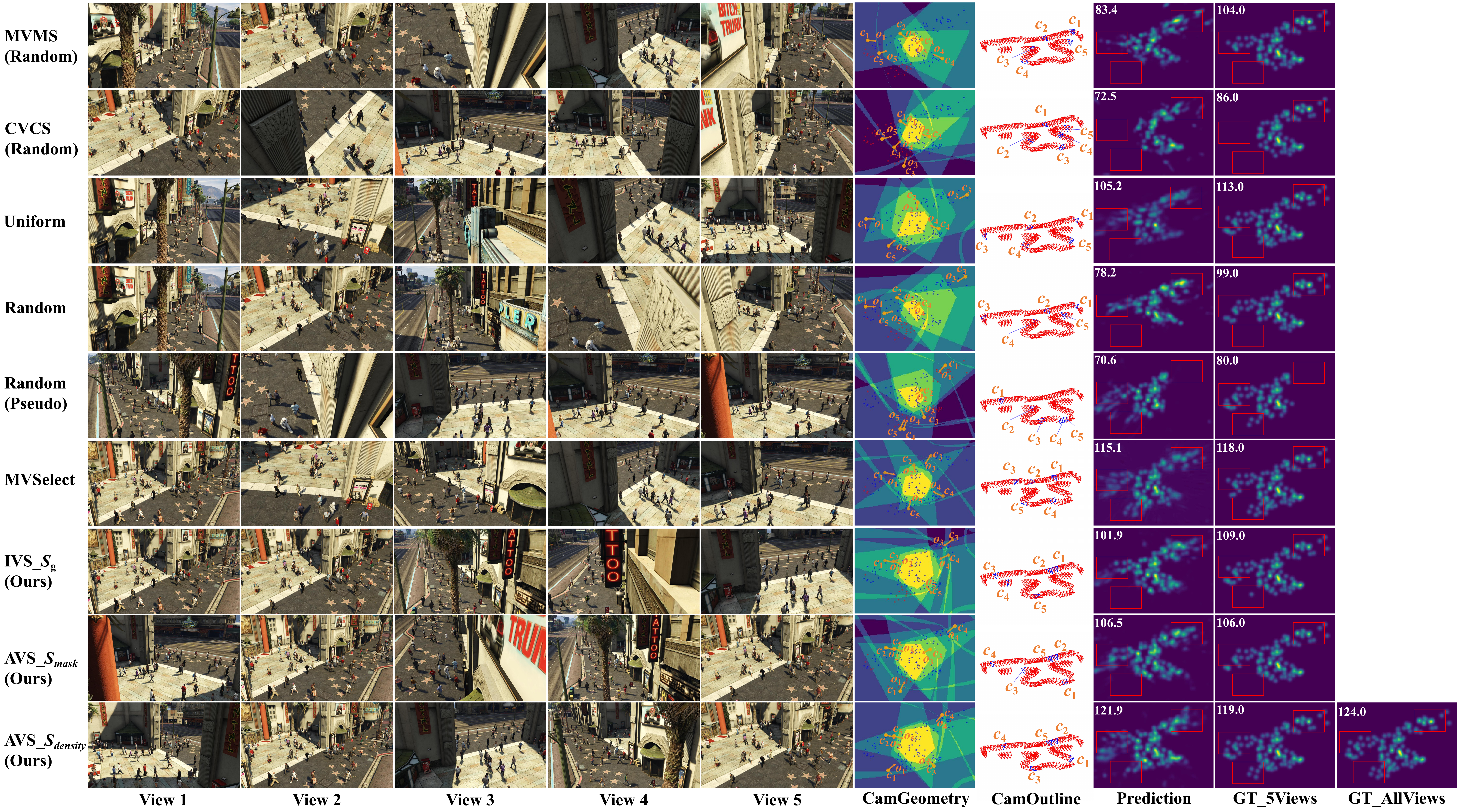}
   \caption{ 
   The view selection and multi-view counting results on CVCS: our methods select better views covering the whole scene and predict better density maps close to the scene-level crowd GT (GT\_AllViews). 
      \lbreb{Blue camera view indicates the selected view in column `CamOutline'. }
   }
\vspace{-0.4cm}
\label{fig:counting_result_main}
\end{figure*}

\begin{table}[t]
\centering
\small


\caption{Comparison of the multi-view counting results on the CityStreet dataset. 
}
\label{table:countingOnCity}
\begin{tabular}{l|c@{\hspace{0.15cm}}c@{\hspace{0.15cm}}c@{\hspace{0.15cm}}c@{\hspace{0.10cm}}}
\hline
    Method                 & MAE$\downarrow$    & MSE$\downarrow$       & NAE$\downarrow$   &  CoverRate$\downarrow$  \\
\hline

   MVMS~\cite{zhang2019wide} (Random)  & 16.82  	& 20.68& 	0.194 & 0.961 \\
    CVCS~\cite{zhang2021cross} (Random) & 15.14  	& 18.25& 	0.180 &0.948   \\

    Uniform    & 12.91  & 15.20& 0.172                & \textbf{0.983} \\
     	 	
   Random     & 13.48  & 16.47& 0.166    &0.965  \\   
   
   Random (Pseudo) & 11.01 	 	& 13.66& 0.128 & 0.965  \\
MVSelect~\cite{hou2024learning}& 10.39 	 	& 13.04& \textbf{0.116} &0.957  \\
\hline
   
   IVS\_$S_{g}$ (Ours)       & 11.28  & 14.36& 0.128                                  &0.946 \\
   AVS\_$S_{mask}$ (Ours)       & 10.26 	& 12.52& 0.121 	                             &0.946 \\
   AVS\_$S_{density}$ (Ours)       & \textbf{9.80}   & \textbf{11.93}& 0.118 &0.946 \\
\hline
\end{tabular}
\vspace{-0.5cm}
\end{table}

\textbf{Comparison methods.}
For the \textit{multi-view counting task}, we compare the proposed AVS with the IVS baseline, `Random', 
`Uniform',
`Random (Pseudo)', 
and MVSelect.
\begin{compactitem}
    \item 
    \lbnew{
    `Random' randomly selects $K$ views at once, trains on the selected views, and shares the same multi-view counting model architecture as ours.
    }    
    \item 
    `Uniform' uses the same multi-view counting model as ours, shares the same multi-view counting model architecture as ours, but replaces the view selection method with the uniform view sampling from all views. 
    \item 
    \lbnew{
    `Random (Pseudo)' selects views in the `Random' way but adopts pseudo-label training.
    }    
    \item 
    MVSelect~\cite{hou2024learning} is a view selection method based on the RL framework, which needs all the labels for model training.
\end{compactitem}
We also compare with previous SOTAs CVCS~\cite{zhang2021cross} and MVMS~\cite{zhang2019wide} with the same labeling budget using the random selection way, denoted as `CVCS (Random)' and `MVMS (Random)'. 
\lbnew{
Since MVSelect cannot be used in the cross-view cross-scene dataset (CVCS), we simply use the number of the prediction action in DQN greater than the maximum action of each scene, and then use a view mask to guarantee a valid action prediction for each scene.
}

\begin{table}[t]
    \centering
    \small
    
\caption{The ablation study on the terms of the AVS score equation $S_{density}$ on CVCS dataset.}
\label{table:viewSelTerm}
\begin{tabular}{l|ccc}
\hline
    Term                 & MAE$\downarrow$    & MSE$\downarrow$       & NAE$\downarrow$     \\
\hline

   $S_{sc}$    & 16.44 & 21.01 & 0.125 \\
   $S_{sc}*S_{ad}$   & 18.56 & 23.39 & 0.138 \\
   $S_{sc}*S_{vd}$   & 14.77 & 18.31 & 0.108 \\
   All ($S_{density}$)  & \textbf{10.99}  & \textbf{13.57} & \textbf{0.083}  \\
\hline
\end{tabular}
\vspace{-0.5cm}
\end{table}

\lbnew{
For the \textit{multi-view localization task}, we still compare the proposed AVS with the IVS baseline, `Random', `Uniform', `Random (Pseudo)', and MVSelect. Moreover, we also compare with recent SOTAs trained with full labels, such as MVDet~\cite{hou2020multiview}, SHOT~\cite{song2021stacked}, MVDeTr~\cite{hou2021multiview}, 3DROM~\cite{qiu20223d}, and M-MVOT~\cite{zhang2024mahalanobis}. These methods represent strong fully-supervised baselines with diverse modeling paradigms, providing a comprehensive benchmark for evaluation.
Note that MVSelect uses \emph{all} labels for joint view selection and crowd localization model training, while we only need to label the selected $K$ views. This setting significantly reduces annotation cost and better reflects practical deployment scenarios where full multi-view annotations are often unavailable or prohibitively expensive, while still maintaining competitive performance.
}

\begin{figure*}[t]
\centering
   \includegraphics[width=0.75\linewidth]{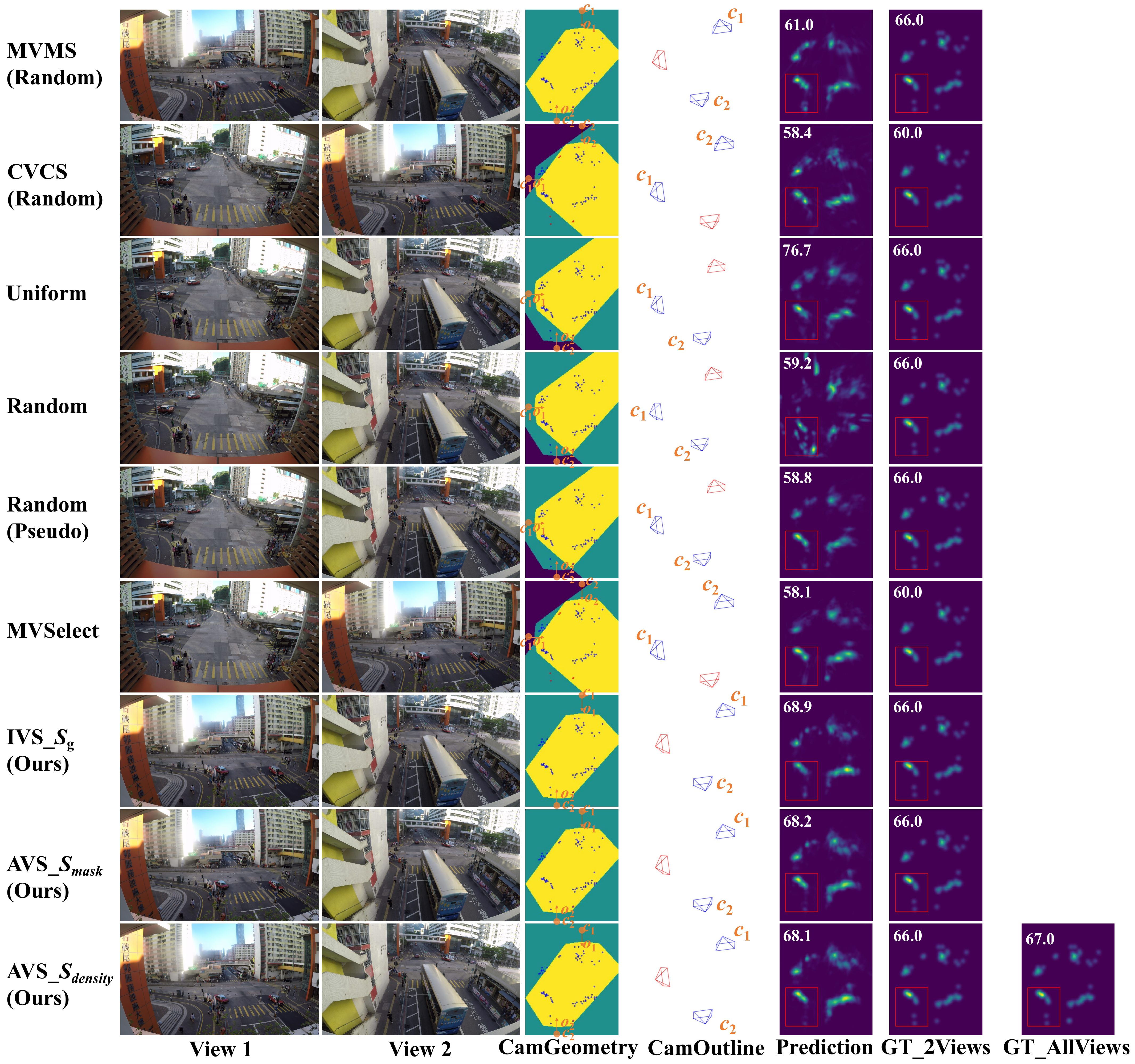}
      \caption{
   \lbnew{
   The view selection and multi-view counting results on CityStreet. Since the CityStreet scene is small, the results of different view selection methods are similar. Our methods still achieve better scene-level prediction results. 
   }
   }
   \vspace{-0.4cm}
\label{fig:counting_city_result_main}
\end{figure*}

\begin{table}[t]
\centering
\small

\caption{The ablation study on $\lambda$ of $S_{density}$ on the CVCS dataset.}
\label{table:viewSelectionLambda}
\begin{tabular}{l|ccc}
\hline
    $\lambda$  & MAE$\downarrow$    & MSE$\downarrow$       & NAE$\downarrow$     \\
\hline

    0.05 & 12.88 & 16.24 &0.096  \\
    0.1  & \textbf{10.99} & \textbf{13.57} & \textbf{0.083} \\
    0.5  & 15.07 & 18.18 &0.112 \\
    1    & 12.74 & 15.87 &0.093 \\
\hline
\end{tabular}

\vspace{8pt}

\caption{The ablation study on the threshold $\tau$ to conduct view addition on the CVCS dataset..}
\label{table:thresholdForViewAddition}
\vspace{4pt}
\begin{tabular}{l|ccc}
\hline
    $\tau$  & MAE$\downarrow$    & MSE$\downarrow$       & NAE$\downarrow$     \\
\hline
15 & 11.47 & 13.99 & 0.086 \\
20 & \textbf{10.99} & \textbf{13.57} & \textbf{0.083}  \\
30 & 12.79 & 15.66 & 0.096  \\
\hline
\end{tabular}

\vspace{8pt}

\caption{The ablation study on the terms of the independent view selection score equation $S_{g}$ on the CVCS dataset.
}
\label{table:viewSelTermIVS}
\begin{tabular}{l|ccc}
\hline
    Term                 & MAE$\downarrow$    & MSE$\downarrow$       & NAE$\downarrow$     \\
\hline

      $S_{sc}$          & 21.65 & 27.46 & 0.159 \\
   $S_{sc} *S_{ad}$   & 19.62 & 25.09 & 0.144 \\
   $S_{sc} *S_{vd}$   & 19.45 & 24.68 & 0.143 \\
   All ($S_{g}$)         & \textbf{14.81}  & \textbf{18.47}  & \textbf{0.110}  \\
\hline
\end{tabular}

\vspace{8pt}

\caption{The ablation study on when to add pseudo label training for AVS\_$S_{density}$ (Ours) on CVCS dataset.
\label{table:pseudoLabels}}
\begin{tabular}{l|ccc}
\hline
    Pseudo               & MAE $\downarrow$    & MSE $\downarrow$       & NAE $\downarrow$     \\
\hline
    None     & 20.17 & 24.77 & 0.156     \\
    ViewSel    & 19.89 & 24.62 & 0.154   \\
    ModelTrain & 11.32 & 14.69 & \textbf{0.083}   \\
    Both (Ours) & \textbf{10.99}  & \textbf{13.57} & \textbf{0.083}  \\
\hline
\end{tabular}
\vspace{-0.5cm}
\end{table}

\textbf{Implementation details.} 
For the multi-view counting task, the input image resolution is 640x360 on CVCS and CityStreet, and a random 160x180 cropping strategy on the scene map is adopted in the training on CVCS. 
\textit{For the view numbers and the values of $K$ and $F$ in each dataset, see Table \ref{tab:cp3_dataset_cost} for details}.
For AVS, during each view expansion iteration, an MAE threshold $\tau$ of 20 is adopted to stop the multi-view counting model training for conducting the next view addition. 
For the multi-view counting model, we use the backbone model in CVCS method with a feature pyramid fusion net (FPN) as the downstream multi-view counting model. 
The model is trained using the SGD optimizer with a learning rate of 1e-3. $\epsilon$ is 1e-10 and $\lambda$ is 0.1 in (\ref{S_vd}), and threshold $\sigma$ is the mean of density map $M_k$.

For the multi-view localization task, during the view selection process, the multi-view crowd localization model training threshold $\tau$ is set to MODA $=$ 40: the model training stops when MODA reaches 40, and then we add the next view. This progressive strategy ensures that each newly introduced view contributes complementary information while avoiding unnecessary training overhead and redundancy among views. Unlike MVSelect, which uses annotations from all views for training, we label only the selected views and apply pseudo-labels to incorporate unlabeled views, thereby reducing annotation cost while maintaining competitive performance.
We use the same MVDet implemented in MVSelect as the downstream multi-view localization model, trained with data augmentation and focal loss in MVDet.
The model is trained using the SGD optimizer, with learning rates of 1e-2 and 5e-2 for Wildtrack and MultiviewX, respectively. $\sigma$ is 0.6 in (\ref{S_mask}), and $\lambda$ and $\epsilon$ are the same as in multi-view counting settings to ensure consistency across tasks.



\begin{table}[t]
\centering
\small

\caption{The ablation on the fixed labeled view number in the pseudo inputs after view selection on the CVCS dataset.}
\label{table:fixedLabeledNumForPseudoInput}
\begin{tabular}{c|ccc}
\hline
    {$\#$ Fixed view }  & MAE$\downarrow$    & MSE$\downarrow$       & NAE$\downarrow$     \\
\hline
0 & 11.07 & 14.27&\textbf{0.083} \\
1 & \textbf{10.99} & \textbf{13.57} & \textbf{0.083}  \\
2 & 11.09 & 13.83&0.084  \\
3 & 12.15 & 15.82&0.091  \\
4 & 12.10 & 14.91&0.090  \\
\hline
\end{tabular} 

\vspace{8pt}

\caption{The ablation on the ratio between labeled inputs and pseudo inputs after view selection on the CVCS dataset.}
\label{table:ratioLabeledView}
\begin{tabular}{l|ccc}
\hline
    Ratio  & MAE$\downarrow$    & MSE$\downarrow$       & NAE$\downarrow$     \\
\hline
    Only Labeled & 12.65 & 15.35 &  0.096 \\
    1:1  & 10.99 & 13.57 &0.083 \\
    1:2  & \textbf{10.85} & \textbf{13.56} & \textbf{0.080} \\
\hline
\end{tabular}

\vspace{8pt}

\caption{The ablation study on the selected view/frame number $K$ (keep $F\!=\!20$)/$F$ (keep $K\!=\!5$) on CVCS dataset.\label{table:viewFrameNum}}
\begin{tabular}{l|ccc|l|ccc}

\hline
    $K$  & MAE  & MSE  & NAE    & $F$ & MAE & MSE    & NAE  \\
\hline
   3 & 15.95  & 20.13 & 0.118   & 5 & 19.01 & 22.91 & 0.144\\
   5 & 10.99  & 13.57 & 0.083   & 10 & 11.69 & 14.56 & 0.088\\
   7 & 10.57  & 13.04 & 0.079   & 20 & 10.99  & 13.57 & 0.083\\
   9 & \textbf{9.82} & \textbf{12.24} & \textbf{0.072} &  40 & \textbf{10.31} & \textbf{12.87} & \textbf{0.077} \\
\hline
\end{tabular}

\vspace{8pt}

\caption{The ablation study on the test of different selected view numbers $K$ using the AVS\_$S_{density}$ model trained with $F=20$ and $K=5$ on the CVCS dataset.}
\label{table:viewNumUsingV5Ckpt}
\begin{tabular}{l|ccc}
\hline
    $K$  & MAE$\downarrow$  & MSE$\downarrow$  & NAE$\downarrow$     \\
\hline
    3 & 15.06 & 18.88 & 0.112 \\
    5 & 10.99 & 13.57 & 0.083 \\
    7 & \textbf{10.53} & \textbf{13.09} & \textbf{0.080} \\
\hline
\end{tabular}
\vspace{-0.5cm}
\end{table}  

\textbf{Evaluation metrics.}
For the multi-view counting task, 
We use mean absolute error (MAE), root mean squared error (MSE), and normalized absolute error (NAE) of the predicted crowd count and the \emph{scene-level} ground-truth count (all crowds in the scene) as metrics: 
    \begin{align}
        MAE &=\frac{1}{N}\sum_{i=1}^{N}{\left| y^{gt}_{i}-\hat{y}_{i} \right|}, \\
        MSE &=\sqrt{\frac{1}{N}\sum_{i=1}^{N}{\left| y^{gt}_{i}-\hat{y}_{i} \right|^{2}}},\\
        NAE &=\frac{1}{N}\sum_{i=1}^{N}{\frac{\left| y^{gt}_{i}-\hat{y}_{i} \right|}{y_i^{gt}}}, 
    \end{align}
where $N$ is the number of the samples, and $\hat{y_i}$ and $y^{gt}_i$ are the predicted count and the corresponding ground truth (GT) count of the $i$-th sample, respectively.
In evaluation, the GT count refers to the crowd number in all views, not the crowd count of the selected views, to indicate the scene-level counting performance. 
Thus, the metrics not only assess the performance of the counting model but also reflect whether the selected views can adequately cover all crowds.
\lb{
The percentage of the crowds covered by the selected views among all crowds in the scene is used to evaluate different view selection methods, and is denoted as `CoverRate':
}
    \begin{align}
        CoverRate &=\frac{1}{N}\sum_{i=1}^{N}{\frac{y^{gt5}_{i}}{y_i^{gt}}}, 
    \end{align}
where $y^{gt5}_{i}$ and  $y_{i}^{gt}$ denote the crowd number in the selected views and all views, respectively. `CoverRate' could indicate the crowd coverage performance of the selected views. 

For the multi-view localization task, we use Multiple Object Detection Accuracy (MODA, MA.), Multiple Object Detection Precision (MODP, MP.),
Precision (P), Recall (R), and F1\_score (F1) as metrics. 
Specifically, $ MODA \! = \!1 \!- \!(FP\!+\!FN\!)\!/\!(\!TP\!+\!FN\!)$ measures the overall performance.
$MODP \!= \!(\sum(1-d[d<t]/t))/TP$ measures the localization precision, where $d$ is the distance from a detected person point to its ground truth and $t$ is the distance threshold set to 0.5m as in \cite{hou2020multiview}. $P = TP/(FP+TP)$, $R = TP/(TP+FN)$, and $F1 = 2P*R/(P+R)$.
$TP$, $FP$, and $FN$ are the number of true positives, false positives, and false negatives, respectively.
\subsection{Multi-view Crowd Counting Results}
We compare the proposed AVS with the IVS baseline and other comparison methods in Table \ref{table:CVCS_results} on the CVCS dataset, and AVS achieves the best performance.
`MVMS (Random)' and `CVCS (Random)'  achieve much worse results because they input random views without good view selection for scene-level counting.
Since the efficiency of the pseudo label training, the results of `Random (Pseudo)' are better than `Random'.
\lbreb{
Compared to `Random' view selection, `Uniform' achieves better performance. But our methods consider view/scene geometries and interaction with downstream models, achieving the best results.
}
\lbnew{
Compared to MVSelect, which uses all labels for training, our methods still achieve better results, further demonstrating the effectiveness of the proposed methods.
}
\textbf{Compared to IVS}, AVS is much better, either with $S_{mask}$ or $S_{density}$.
Even though IVS$\_S_g$ has a close CoverRate to AVS\_$S_{density}$, its scene-level counting performance is much worse than AVS\_$S_{density}$. This shows the superiority of AVS, which optimizes the view selection and the multi-view counting model training jointly for better scene-level results. 
$S_{density}$ is better than $S_{mask}$ because $S_{density}$ considers the crowd density levels as well as the location information in view selection, while $S_{mask}$ only utilizes the location information.
Note that the CVCS dataset is a multi-scene dataset, and our methods could perform \textit{cross-scene training and testing}, demonstrating the flexibility and generalization ability.
\lbnew{
We also compare the proposed view selection methods with the comparison methods on CityStreet in Table \ref{table:countingOnCity}. The table shows that results with view selection methods are better than those without. However, the proposed methods overall achieve the best results, further indicating that equipping a well-designed view selection method is essential for scene-level tasks with limited labels.
}



The \textbf{visualizations} on CVCS are shown in Fig.~\ref{fig:counting_result_main}, where the inputs are variable for different methods. `GT\_$K$Views' and `GT\_AllViews' are the GT constructed from the $K$ selected views and all views. The former is used for training, and the latter is for evaluation. 
It's observed that the `GT\_$K$Views' of our method `AVS\_$S_{density}$' contains the most crowds, 
and our method can also cover more crowds in the scene (red dots in `CamGeometry' indicate crowds not covered by the selected views), indicating the efficacy of our view selection method. 
The predictions further highlight the advantages of our method, whereas the comparison methods fail to capture the regions highlighted by the red boxes. 
\lbnew{
Fig.~\ref{fig:counting_city_result_main} shows the visualizations on CityStreet. Because the CityStreet scene is small, the GT between selected views and the scene is similar. The figure shows that our methods can select more adaptive views, enabling the counting model to achieve better scene-level performance, thereby demonstrating the advantages of the proposed methods.
}





\begin{table}[t]
\centering
\small

\vspace{8pt}  

\captionof{table}{The ablation study on the multi-view counting models using AVS\_$S_{density}$ on the CVCS dataset.
}
\label{table:backbone}
\begin{tabular}{l|ccc}
\hline
    Model                & MAE $\downarrow$    & MSE $\downarrow$       & NAE $\downarrow$     \\
\hline
   Backbone      & 11.20 & 14.70 & 0.083 \\

   Backbone+CVCS & 15.25 & 18.63 & 0.113 \\
   Backbone+MVMS & \textbf{9.74}  & \textbf{13.31} & \textbf{0.074} \\
   Backbone+FPN  & 10.99  & 13.57 & 0.083  \\

\hline
\end{tabular}

\vspace{4pt}  

\captionof{table}{The ablation study on the frame number used 
for view selection when testing AVS\_$S_{density}$ on the CVCS dataset. Note that the counting model still tests on all frames (100) for the result report.}
\label{table:testDiffFrameUsingV5Ckpt}
\vspace{4pt}  
\begin{tabular}{l|cccc}
\hline
  Frames  & MAE$\downarrow$  & MSE$\downarrow$  & NAE$\downarrow$ & Time (h)$\downarrow$    \\
\hline
  5   & 11.43 & 14.20 &0.085 &\textbf{2.0} \\
  20  & 10.97 & 13.60 &\textbf{0.082} &7.1 \\
50  & \textbf{10.91} & \textbf{13.54} & \textbf{0.082} &15.6\\    
  100  &10.99 & 13.57 &0.083 &30.7 \\  	
\hline
\end{tabular}

\vspace{-0.5cm}
\end{table}

\subsection{Ablation Study on Multi-view Crowd Counting}

\lbnew{
To better validate the efficiency of our proposed methods, the experiments are mainly conducted on CVCS dataset. The results are as follows.
}

\textbf{Ablation on the terms of $S_{density}$.}
We perform ablation studies on the usage of the 3 terms in $S_{density}$ in Table~\ref{table:viewSelTerm}: using $S_{sc}$, using $S_{sc}*S_{ad}$ or $S_{sc}*S_{vd}$, or using all 3 terms ($S_{density}$). 
Compared to only using $S_{sc}$, adding $S_{vd}$ improves the results, while adding $S_{ad}$ without the view diversity term $S_{vd}$ achieves worse results, due to selected views being placed at the similar locations and directions, reducing the multi-view fusion performance (as shown in Fig.~\ref{fig:viewSel} (b)).
Using all 3 terms is the best because it can select views covering most of the crowd with a larger overlapping area for better multi-view fusion (see Fig.~\ref{fig:viewSel} (d)),  
which indicates each term's contribution to the final view selection performance.

\textbf{Ablation study on $\lambda$ in $S_{density}$.} 
We conduct experiments on the hyper-parameter $\lambda$ in $S_{density}$ to evaluate the sensitivity of the term $S_{vd}$.
\lbnew{
As shown in Table~\ref{table:viewSelectionLambda}, when $\lambda$ is too small, the view diversity constraint becomes weak, which may lead to suboptimal view sets (see Fig.~\ref{fig:viewSel} (a) and (b)). In contrast, excessively large $\lambda$ values ($0.5, 1$) overemphasize view diversity, diminishing the contributions of $S_{sc}$ and $S_{ad}$ and resulting in inferior performance. Therefore, compared with other settings, $\lambda=0.1$, which provides a balanced sensitivity, is more suitable for the proposed scoring function.
}

\textbf{Ablation study on the threshold $\tau$.} 
We experiment with the downstream model performance threshold $\tau$ for view addition. As shown in Table \ref{table:thresholdForViewAddition}, $\tau=20$ is more suitable for our AVS framework.
When $\tau$ is too large, the counting model is under training, which may result in bad final performance;
When $\tau$ is too small, the counting model may overfit at each step of adding views, leading to poor final performance. $\tau=20$ achieves a balance between the current and final counting model performance.

\textbf{Ablation study on the terms of $S_g$.}
In addition to the term ablation study for $S_{density}$ of AVS framework in the manuscript, we conduct additional ablation experiments on the 3 terms in $S_{g}$ of the IVS framework in Table \ref{table:viewSelTermIVS}: using $S_{sc}$, $S_{sc}*S_{ad}$, $S_{sc}*S_{vd}$, or using all 3 terms (namely $S_{g}$). 
The results are similar to the experiments on $S_{density}$ and demonstrate that each term in $S_{g}$ contributes to the final performance by leveraging the scene and view geometries. 

\begin{table}[t]
\centering
\small

\caption{The ablation studies on the random frame selection method on the CVCS dataset ($F=20$).}
\label{table:frameSel_CVCS}
\begin{tabular}{l|ccc}
\hline
    Frame Select  & MAE$\downarrow$    & MSE$\downarrow$       & NAE$\downarrow$     \\
\hline
    Random/Uniform        & 12.28 & 15.46 & 0.092\\
    AVS (Ours)    & \textbf{10.99}  & \textbf{13.57} & \textbf{0.083} \\
\hline
\end{tabular}
\end{table}

\begin{figure}[t]
\centering
   \includegraphics[width=\linewidth]{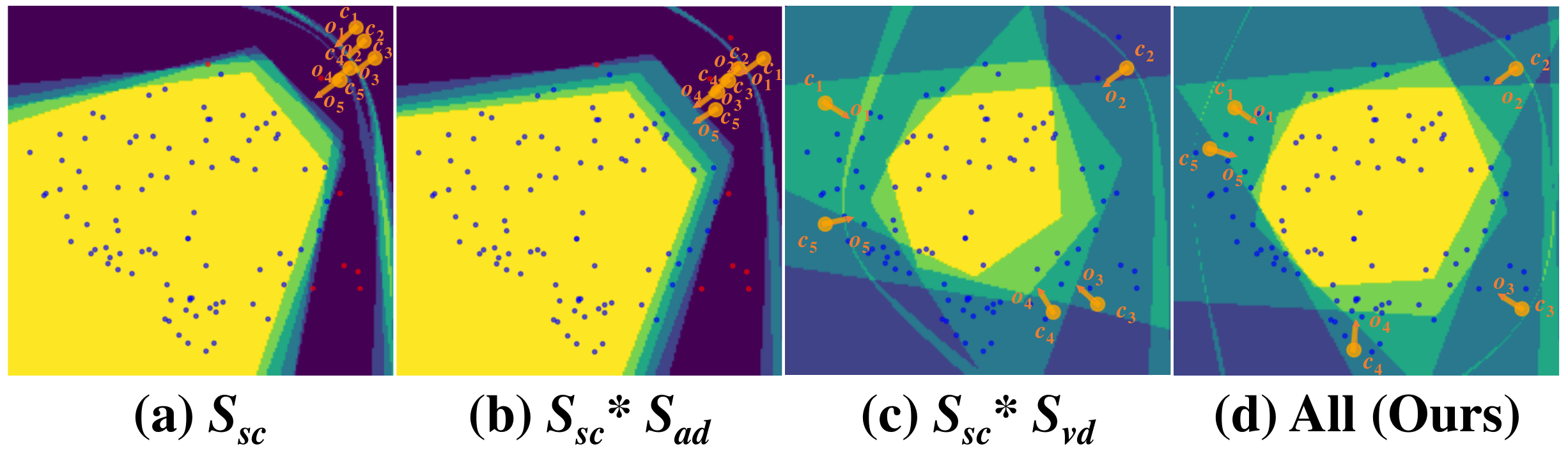}
   \caption{ 
   The selected view positions ($c_j$) and directions ($o_j$) for
   different terms. Red dots are uncovered crowds.
   }
\vspace{-0.6cm}
\label{fig:viewSel}
\end{figure}

\begin{figure*}[t]
\centering
   \includegraphics[width=0.9\linewidth]{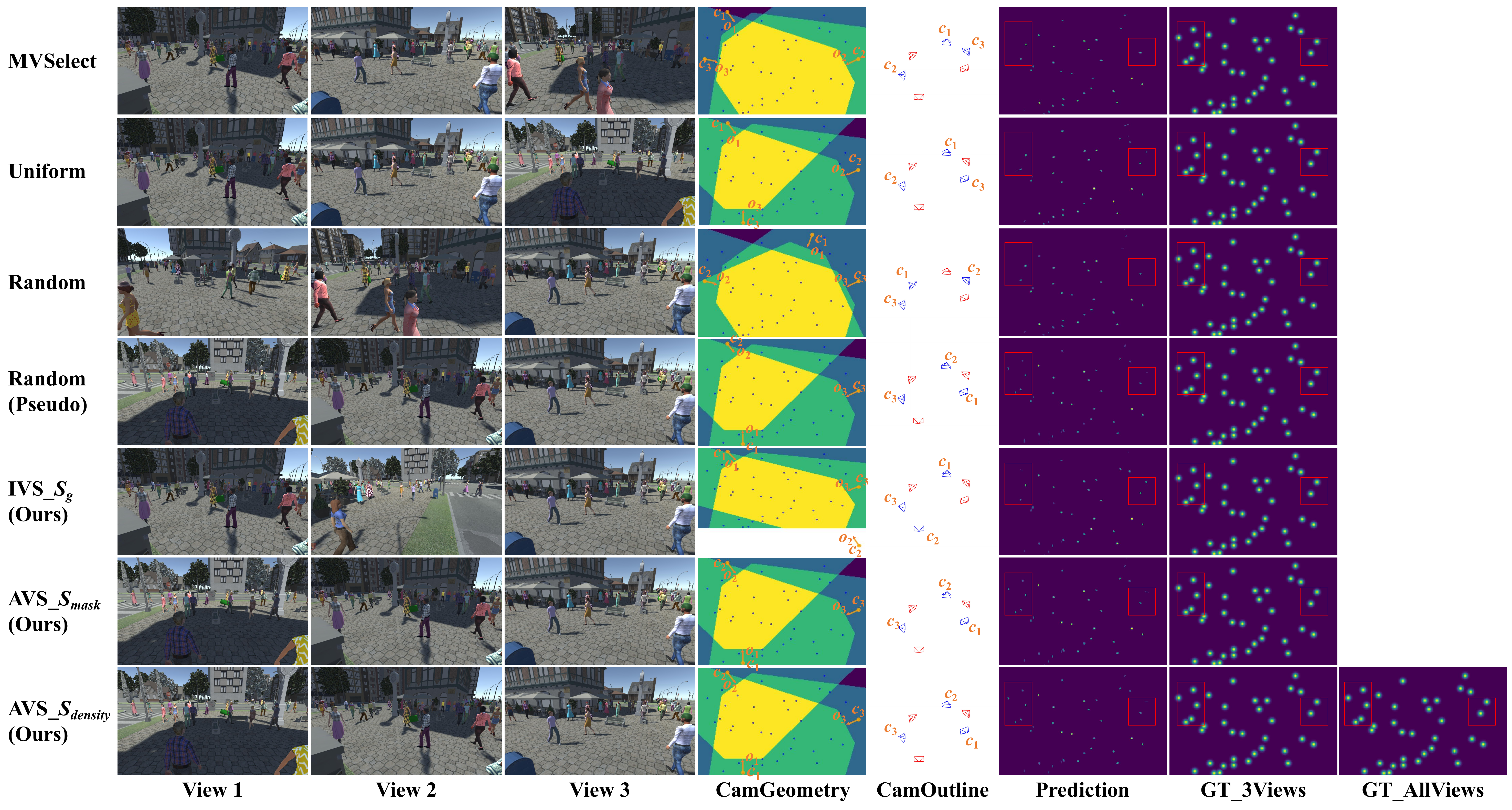}
   \caption{
   The view selection and multi-view localization results on MultiviewX. Our methods select views that achieve better prediction results.
   \lbreb{Blue camera view indicates the selected view in column `CamOutline'. }
   }
\vspace{-0.4cm}
\label{fig:multiviewx_result}
\end{figure*}

\textbf{Ablation study on pseudo labels.} 
The ablation studies on the pseudo labels for AVS\_$S_{density}$ are shown in Table \ref{table:pseudoLabels}.
We compare the method without using the pseudo labels in the model training (None), using pseudo labels only at the view selection stage (ViewSel), using pseudo labels only at the final model training stage when view selection is finished (ModelTrain), or using pseudo labels at both stages (Ours). The results show that pseudo labels can indeed improve the performance when added at any stage, and adding pseudo labels at both stages can obtain the best performance.

\textbf{Ablation study on the number of labeled views for pseudo inputs.}
We conduct experiments on the number of labeled views in pseudo inputs after view addition. Specifically, we vary the proportion of labeled views used to construct pseudo inputs to analyze their impact on model performance. With more labeled views in pseudo inputs, the randomness of the pseudo inputs is reduced, leading to less diverse training signals and consequently worse generalization ability, as shown in Table \ref{table:fixedLabeledNumForPseudoInput}. This observation suggests that excessive reliance on labeled views may cause the model to overfit to limited view configurations.
Hence, we only retain one labeled view in pseudo inputs, which preserves sufficient randomness and diversity, achieving better performance and improved robustness across different view combinations.

\textbf{Ablation study on the ratio pseudo input.} 
We conduct experiments on the ratio between pseudo inputs and labeled inputs. Generally speaking, using more pseudo inputs can improve the model's robustness. However, because the labels are imperfect, we cannot rely solely on the pseudo inputs for model training. As shown in Table \ref{table:ratioLabeledView}, the ratio of 1:2 between labeled inputs and pseudo inputs achieves the best result, but we use the 1:1 ratio between labeled inputs and pseudo inputs for time tradeoff.

\textbf{Different view and frame number.
}
We conduct ablation studies on the selected view number $K$ and frame number $F$ for AVS\_$S_{density}$ in Table \ref{table:viewFrameNum}, with other settings kept the same (except $\tau\!=\!30$ for 5 frames for its poor performance).
As $K$ increases, more views are provided to cover the whole scene, generally achieving better scene-level counting performance.
With more frames, the multi-view counting model is trained on more labeled data, yielding better results.

\begin{table}[t]
\centering
\small
\caption{
\lbreb{The ablation on the comparison method with uniform view selection and pseudo-labels on CVCS dataset.}
}
\label{table:table_ablation_comparison_uniform}
\begin{tabular}{l@{\hspace{0.10cm}}|c@{\hspace{0.1cm}}c@{\hspace{0.12cm}}c@{\hspace{0.1cm}}|c@{\hspace{0.1cm}}}
\hline
    Method                  & MAE $\downarrow$    & MSE  $\downarrow$      & NAE  $\downarrow$  & CoverRate $\uparrow$  \\
\hline

MVMS (Random)               & 37.22 &	44.74 &	0.276 &	0.872 \\
MVMS (Random, Pseu.)        & 31.11 &	38.29 &	0.229 &	0.872 \\
MVMS (Uniform)              & 30.38 &	36.42 &	0.225 &	0.945 \\
MVMS (Uniform, Pseu.)       & 23.37 &	28.49 &	0.173 &	0.945 \\
CVCS (Random)               & 30.97 &	36.47 &	0.228 &	0.901 \\
CVCS (Random, Pseu.)        & 24.89 &	30.54 &	0.182 &	0.901 \\
CVCS (Uniform)              & 28.06 &	32.91 &	0.209 &	0.945 \\
CVCS (Uniform, Pseu.)       & 22.28 &	27.97 &	0.164 &	0.945 \\

    Random                                                  & 36.59 & 42.06 & 0.271 &0.885 \\
    Random (Pseu.)                                         & 28.22 & 33.73 & 0.208 &0.885 \\  
    Uniform
    & 21.76 & 25.75 & 0.163 &0.945 \\
    Uniform (Pseu.)  
   & 15.69 & 19.92 & 0.115 &0.945 \\
\hline
    IVS\_$S_g$ (Baseline, Ours) & {14.81}  & {18.47}  & {0.110}  & 0.954\\
    AVS\_$S_{mask}$ (Ours)    & 12.53 & 15.33 & 0.093                              &0.955          \\
    AVS\_$S_{density}$ (Ours) & \textbf{10.99}  & \textbf{13.57} & \textbf{0.083}  &\textbf{0.960} \\
\hline
\end{tabular}

\vspace{-0.6cm}

\end{table}

\textbf{Ablation study on the testing view number.}
We conduct extra experiments on the testing with different view numbers $K$ (3, 5, and 7) using the AVS\_$S_{density}$ model trained with $K=5$ views. As shown in Table \ref{table:viewNumUsingV5Ckpt}, with the view number increasing, the model's performance improves relatively, as more regions are covered, demonstrating the good generalization ability of the proposed AVS framework to variable numbers of input views.

\textbf{Ablation on the different multi-view counting models.}
The ablation studies on the multi-view counting models in the active view selection framework AVS\_$S_{density}$ are shown in Table \ref{table:backbone}. The `Backbone' model is the same backbone model of CVCS~\cite{zhang2021cross}. `+MVMS',`+CVCS', and `+FPN' mean adding the multi-view multi-scale selection~\cite{zhang2019wide}, camera view selection~\cite{zhang2021cross}, and the feature pyramid fusion module~\cite{lin2017feature} to the `Backbone' model, respectively. From the table, using `Backbone+MVMS' achieves the best multi-view counting results. `Backbone+CVCS' achieves worse results than `Backbone', perhaps due to the requirement of more labeled data to learn a good camera selection module, whereas our task setting provides limited labeled data. We adopt the `Backbone+FPN' model as the multi-view counting model in our experiments to balance training efficiency and performance.

\begin{figure*}[t]
\centering
   \includegraphics[width=\linewidth]{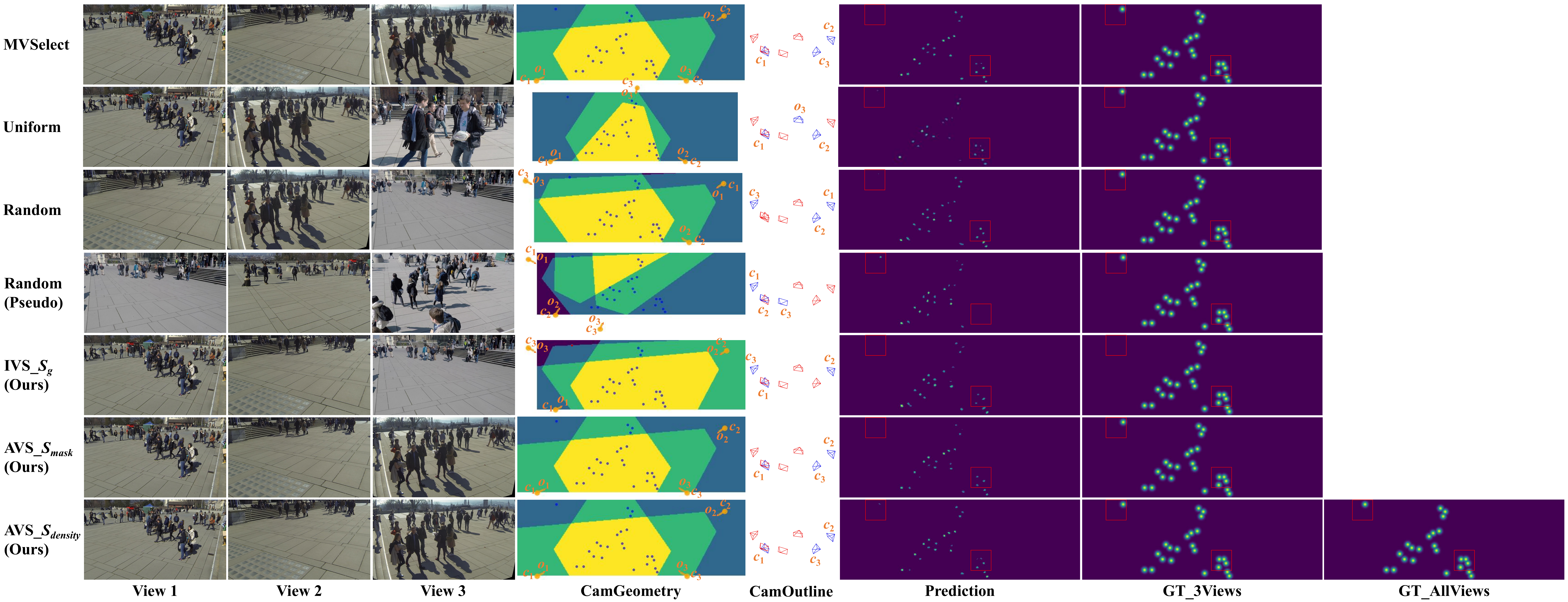}
   \caption{ 
   The view selection and multi-view localization results on Wildtrack. Our methods still select views that achieve better prediction results, further demonstrating the efficiency of the proposed method.
   }
\vspace{-0.2cm}
\label{fig:wildtrack_result}
\end{figure*}

\begin{table*}[t]

\caption{Comparison of the multi-view localization results on Wildtrack and MultiviewX. AVS achieves the best performance among all partial-labeled methods (3 views) and outperforms MVSelect trained with the ground truth of full labels (all views). Bold indicates the best result among all partial-labeled methods.
}
\small
\centering
\begin{tabular}{ll|ccccc|ccccc}

\hline
        & Dataset &  \multicolumn{5}{c|}{MultiviewX}  &  \multicolumn{5}{c}{Wildtrack}  \\
Label   & Method         & MODA$\uparrow$    & MODP$\uparrow$        & P$\uparrow$        & R$\uparrow$     & F1$\uparrow$
                     & MODA$\uparrow$    & MODP$\uparrow$        & P$\uparrow$        & R$\uparrow$     & F1$\uparrow$ \\
\hline
\multirow{6}{*}{Full}
    & MVDet   ~\cite{hou2020multiview}          & 83.9    & 79.6        & 96.8        & 86.7  & 91.5     & 88.2    & 75.7        & 94.7        & 93.6  & 94.1
                                                  \\
    & SHOT    ~\cite{song2021stacked}           & 88.3    & 82.0        & 96.6        & 91.5  & 94.0       & 90.2    & 76.5        & 96.1        & 94.0  & 95.0
                                                  \\
    & MVDeTr  ~\cite{hou2021multiview}         & 93.7    & 91.3        & 99.5        & 94.2  & 97.8     & 91.5    & 82.1        & 97.4        & 94.0  & 95.7        \\
    & 3DROM   ~\cite{qiu20223d}     & 95.0    & 84.9        & 99.0        & 96.1  & 97.5    & 93.5    & 75.9        & 97.2        & 96.2  & 96.7                              \\
    & M-MVOT~\cite{zhang2024mahalanobis} & 96.7 & 86.1& 98.8& 97.9& 98.3& 92.1& 81.3& 94.5& 97.8& 96.1 \\
   & MVSelect ~\cite{hou2024learning}  & 88.1 & 89.8 & 98.2 & 89.7  & 93.8 &  88.6  & 79.9 & 93.3 & 94.2 & 93.7 \\
\hline
\multirow{6}{*}{Partial}
   &  Random                     & 85.3 & 80.8 & 97.3 & 87.7  & 92.2 &  80.6  & 75.8 & 93.0 & 87.1 & 89.8 \\ 
   &  Random (Pseudo)            & 85.5 & 81.1 & 97.5 & 87.7  & 92.4 &  82.8  & 75.4 & 93.8 & 88.5 & 91.0 \\  
   & Uniform           & 82.6 & \textbf{87.3} & 96.4  & 85.8  &  90.8   & 84.6 & \textbf{79.7} & 95.1 & 89.2 & 92.0 \\
    & IVS\_${S_g}$ (Ours) & 87.1 & {83.3} 	 & 97.5 & 89.4 &93.3 & 87.4 &	75.7 &	93.2 	&\textbf{94.2}& 	93.7 \\
   & AVS\_$S_{mask}$ (Ours)   & 87.9 & 80.5 & 97.3 & 90.4  & 93.7 &  87.7  & 77.0 & 95.5 & 92.0 & 93.7 \\
   & AVS\_$S_{density}$ (Ours)& \textbf{89.2} & {82.1} & \textbf{98.0} & \textbf{91.0}  & \textbf{94.4} &  \textbf{89.6}  & 76.7 & \textbf{96.1} & 93.4 & \textbf{94.7} \\
\hline
\end{tabular}

\vspace{-0.5cm}
\label{table:Wildtrack_MultiviewX_results}
\end{table*}

\textbf{Ablation study on the frame number for view selection and computation time in testing on the CVCS dataset.}
We conduct experiments on using different frame numbers in view selection when testing AVS\_$S_{density}$ trained with labeled $F=20$ frames and $K=5$ views. Table \ref{table:testDiffFrameUsingV5Ckpt} indicates that only using 20 frames at testing to conduct view selection can achieve almost similar results compared with using 50 frames or 100 frames, with much less computation time, though. 
This demonstrates that the adopted frame initialization approach can effectively select representative frames for view selection and reduce testing computation time.
Note that the counting model still tests on all frames (100) for the performance report.

\textbf{
Different frame selection methods.}
We conduct an additional random frame selection method, and our proposed frame selection method outperforms the random or uniform frame selection shown in Table \ref{table:frameSel_CVCS}. 
The reason is that our method selects frames across various lighting conditions using cosine similarity, thereby enhancing the sample diversity and the network's robustness.
\lbreb{
For uniform frame sampling, because of unordered frames on CVCS dataset, random frame sampling is equivalent to uniform frame sampling.
}

\textbf{Ablation study on comparison method with uniform view selection and pseudo label.} 
As shown in Table \ref{table:table_ablation_comparison_uniform}, uniform view selection consistently achieves better scene-level performance than random selection, highlighting the importance of a reasonable view selection method. However, our methods further outperform the uniform strategy, indicating that explicitly modeling view/scene geometries as well as crowd density and localization information enables the selection of views that are more suitable for accurate prediction.
In addition, we incorporate pseudo-labels (Pseu.) into the comparison methods. The results show that introducing pseudo-labels consistently improves their performance, demonstrating the effectiveness of this strategy. Nevertheless, our methods still achieve superior results, underscoring that a well-designed, geometry- and density-aware view selection plays a more critical role in overall performance improvement.

\subsection{Multi-view Crowd Localization Results}

As shown in Table \ref{table:Wildtrack_MultiviewX_results}, we compare our methods with other view-selection-based methods (MVSelect, `Random', and  `Random (Pseudo)', `Uniform') and full-label supervised methods.
The results show that AVS\_$S_{density}$ outperforms the random view selection methods `Random', `Random (Pseudo)', `Uniform', and MVSelect, demonstrating the advantages of using joint 
optimization
of the view selection and downstream model training. 
Compared to IVS, 
AVS
achieves better performance, either with  $S_{mask}$ or $S_{density}$.
$S_{density}$ is better than $S_{mask}$, which also proves AVS's effectiveness due to considering both crowd density-level information and location information, and the view/scene geometries in the view selection.
Compared to MVDet, SHOT, MVDeTr, 3DROM, and M-MVOT, which are trained on all input views and labels, the proposed active view selection framework ($S_{density}$) outperforms MVDet on both MultiviewX and Wildtrack with limited label costs, also proving the advantages of our methods.
Note that MVSelect also relies on all camera view labels (annotations and calibrations) in the model training and cannot perform on novel new scenes with different views and scene settings, while \textit{our methods only rely on limited view labels with wider application scenarios (as on CVCS)}.





We visualize the results from the comparison methods and the proposed methods on MultiviewX and Wildtrack in Fig.~\ref{fig:multiviewx_result} and \ref{fig:wildtrack_result}. 
Because both datasets are smaller than CVCS, a few views can easily cover the crowd on the ground. Our method achieves better scene-level localization performance than the comparisons, as shown in the red box regions, where our methods have fewer missing points.
Furthermore, our AVS framework jointly optimizes view selection and multi-view localization model, and employs novel pseudo-labels during model training to improve localization performance.

\subsection{Ablation Study on Multi-view Crowd Localization}

\textbf{Ablation study on pseudo labels.}
The ablation studies on the pseudo labels for the active view selection framework ($S_{density}$) are shown in Table \ref{table:PseudoLabels2}: no pseudo labels (None), adding at view selection (ViewSel) or final model training stage (ModelTrain) after view selection, or both (Ours).
Similarly, we add the pseudo-label training at different stages and compare their influence on the performance.
The results show that regardless of which stage, pseudo labels can improve the performance. 
On Wildtrack, adding pseudo labels is more effective at the view selection stage. The possible reason is that the view difference is larger in Wildtrack, and thus the pseudo-label training is more useful for the model to generalize to new views. Anyway, adding pseudo-label training at both stages can achieve the best performance. 


\begin{table}[t]

\small
\centering

\caption{The ablation study on the pseudo labels.}
\label{table:PseudoLabels2}
\begin{tabular}{@{\hspace{0.02cm}}l@{\hspace{0.05cm}}|c@{\hspace{0.08cm}}c@{\hspace{0.08cm}}c@{\hspace{0.08cm}}c@{\hspace{0.08cm}}c@{\hspace{0.08cm}}|c@{\hspace{0.08cm}}c@{\hspace{0.08cm}}c@{\hspace{0.08cm}}c@{\hspace{0.08cm}}c@{\hspace{0.02cm}}}
\hline
    Dataset &  \multicolumn{5}{c|}{MultiviewX}  & \multicolumn{5}{c}{Wildtrack}\\
    Pseudo        & MA.    & MP.       & P      & R  & F1   & MA.    & MP.       & P     & R & F1 \\
\hline
   None        & 86.2 & 73.0 & \textbf{98.4} & 87.6 & 92.7 & 79.6 & \textbf{77.6} & 94.2 & 84.9 &89.3\\
   ViewSel     & 86.9 & 79.8 & 97.6 & 89.1 & 93.1 & 83.6 & 75.9 & 94.7 & 88.6 &91.5\\
   ModelTrain   & 87.8 & 81.6 & 98.0 & 89.7 & 93.6 & 79.9 & 77.5 & 95.8 & 83.6 &89.3\\
   Both (Ours)  & \textbf{89.2} & \textbf{82.1} & 98.0 & \textbf{91.0} & \textbf{94.4} & \textbf{89.6} & 76.7 & \textbf{96.1} & \textbf{93.4} &\textbf{94.7} \\
\hline
\end{tabular}

\vspace{8pt}  

\caption{
\lbreb{
The ablation study on the threshold $\tau$ to conduct view addition on the MultiviewX and Wildtrack.
}
}
\label{table:supple_thresholdTau_localization}
\begin{tabular}{c@{\hspace{0.0cm}}|c@{\hspace{0.10cm}}c@{\hspace{0.12cm}}c@{\hspace{0.12cm}}c@{\hspace{0.12cm}}c@{\hspace{0.12cm}}|c@{\hspace{0.12cm}}c@{\hspace{0.12cm}}c@{\hspace{0.12cm}}c@{\hspace{0.12cm}}c@{\hspace{0.12cm}}}
\hline
    Dataset     & \multicolumn{5}{c|}{MultiviewX} & \multicolumn{5}{c}{Wildtrack} \\
    $\tau$        & MA.    & MP.       & P      & R  & F1    & MA.    & MP.       & P      & R    & F1  \\
\hline


30 & 87.5 & 81.3 & 97.3 & 90.0 & 93.5 & 85.3 & 76.3 & 96.0 & 89.0 & 92.4 \\
40 & \textbf{89.2} & \textbf{82.1} & 98.0 & \textbf{91.0} & \textbf{94.4} & \textbf{89.6} & \textbf{76.7} & 96.1 & \textbf{93.4} & \textbf{94.7} \\
50 & 87.4 & 78.3 & \textbf{98.1} & 89.1 & 93.4 & 87.4 & 76.1 & \textbf{96.4} & 90.8 & 93.5 \\
60 & 87.0 & \textbf{82.1}& 97.9 & 88.9 & 93.2 & 86.9 & 73.6 & 96.0 & 90.7 & 93.2 \\

\hline
\end{tabular}
\vspace{-0.4cm}
\end{table}

\textbf{Ablation study on threshold $\tau$.} As a hyperparameter, the threshold $\tau$ controls when  to select the next view. 
For the multi-view crowd localization task on Wildtrack and MultiviewX, AVS framework selects the next view when the MODA during training exceeds the threshold $\tau$. 
AVS framework jointly optimizes view selection and downstream task models. 
Hence, a well-trained model can produce a more accurate density map, thereby reducing prediction errors during view selection. Consequently, we need an appropriate model to select views by controlling the threshold $\tau$. As shown in Table \ref{table:supple_thresholdTau_localization}, threshold $\tau=40$ is more suitable on Wildtrack and MultiviewX, achieving the better scene-level performance.

\textbf{Comparison of training with different frame numbers.}
We compare the proposed AVS\_{$S_{density}$} and MVSelect by training with different numbers of frames (36, 72, 180, and 360) on 
MultiviewX in Table \ref{table:frameNumMultiviewX}. 
The testing set is the same (40 frames).
As the number of training frames increases, performance improves, and our method outperforms MVSelect across a range of frame counts, demonstrating its efficiency relative to MVSelect.

\begin{table}
\centering
\small

\caption{The ablation study on the training frame number $F$ on the MultiviewX dataset.}
\label{table:frameNumMultiviewX}
\begin{tabular}{l@{\hspace{0.0cm}}|c@{\hspace{0.10cm}}c@{\hspace{0.12cm}}c@{\hspace{0.12cm}}c@{\hspace{0.12cm}}c@{\hspace{0.12cm}}|c@{\hspace{0.12cm}}c@{\hspace{0.12cm}}c@{\hspace{0.12cm}}c@{\hspace{0.12cm}}c@{\hspace{0.12cm}}}
\hline
    Method     & \multicolumn{5}{c|}{AVS\_$S_{density}$} & \multicolumn{5}{c}{MVSelect} \\
    Frame        & MA.    & MP.       & P      & R  & F1    & MA.    & MP.       & P      & R    & F1  \\
\hline
   36  & 73.6 & 76.4 & 95.8 & 77.0 & 85.4 												 & 61.5 & 73.4 & 96.9 & 63.5 & 76.7 \\
   72  & 81.2 & 71.6 & 95.9 & 84.8 & 90.0 												 & 69.8 & 51.2 & 96.6 & 72.4 & 82.8 \\
   180 & 86.0 & 80.8 & 96.7 & 89.1 & 92.7 												 & 77.6 & 60.0 & 97.1 & 80.0 & 87.7 \\
   360 & \textbf{89.2} & \textbf{82.1} & \textbf{98.0} & \textbf{91.0} & \textbf{94.4}   & \textbf{88.1} & \textbf{89.8} & \textbf{98.2} & \textbf{89.7}  & \textbf{93.8} \\
\hline
\end{tabular}

\vspace{8pt}  

\caption{The ablation study on the selected view number $K$ on the MultiviewX dataset, training and testing using the same $K$ views.}
\label{table:viewNumMultiviewX}
\begin{tabular}{c@{\hspace{0.0cm}}|c@{\hspace{0.10cm}}c@{\hspace{0.12cm}}c@{\hspace{0.12cm}}c@{\hspace{0.12cm}}c@{\hspace{0.12cm}}|c@{\hspace{0.12cm}}c@{\hspace{0.12cm}}c@{\hspace{0.12cm}}c@{\hspace{0.12cm}}c@{\hspace{0.12cm}}}
\hline
    Method     & \multicolumn{5}{c|}{AVS\_$S_{density}$} & \multicolumn{5}{c}{MVSelect} \\
    View        & MA.    & MP.       & P      & R       & F1    & MA.    & MP.       & P      & R    & F1  \\
\hline


   2 & 81.7 		 & 80.4 		 & 97.1 		 & 84.3 		 & 90.2 			 & 77.5 &	82.9 &	97.4 &	79.6 & 	87.6  \\
   3 & 89.2 		 & \textbf{82.1} & 98.0 		 & 91.0 		 & 94.4 			 & 88.1 &	\textbf{89.8} &	\textbf{98.2} &	89.7 & 	93.8  \\
   4 & 92.0 		 & 78.7 		 & 98.2 		 & 93.7 		 & 95.9 			 & 89.6 &	87.9 &	97.5 &	92.0 & 	94.7  \\
   5 & \textbf{93.4} & 79.2 		 & \textbf{98.3} & \textbf{95.1} & \textbf{96.6}     & \textbf{92.3} &	89.1 &	97.8 &	\textbf{94.5}&\textbf{96.1}  \\ 
   
\hline
\end{tabular}
\vspace{-0.4cm}
\end{table}

\textbf{Comparison of training and testing with different view numbers.}
We conduct experiments on the proposed method and MVSelect, training, or testing with different view numbers.
For the results shown in Table \ref{table:viewNumMultiviewX}, the view number used during both the training and testing processes is the same. As the number of views increases, the performance gradually becomes better, and it is nearly converged when $K\geq 3$. Furthermore, regardless of the number of views used, our method still outperforms MVSelect, indicating the advantage of the proposed method.
The results shown in Table \ref{table:viewNumUsingV3CkptMultiviewX} use the model trained on 3 views to test with different view numbers, $K=2, 3, 4, 5$.
Our method generally outperforms MVSelect on all view numbers,
showing the proposed method's generalization ability to different input view numbers. Note that MVSelect utilizes labels from all views for training and is challenging to apply to new scenes due to the reinforcement learning framework.

\textbf{Comparison of frame selection methods.}
Compared with the uniform and random frame selection methods in Table \ref{table:differentFrameSel_wildtrack_multiviewx}, our frame selection approach is still better. As mentioned above, our method selects frames with various lighting conditions using cosine similarity, which can enhance the diversity of the sample and the robustness of the network.


\textbf{Comparison of the model costs.}
As shown in Table \ref{table:training_cost},
due to using pseudo labels and model predictions for view selection, our training time is higher than MVSelect and Random, but comparable to others. 
Yet, our test speed is similar to baselines and faster than MVSelect since no extra reinforcement learning network is needed.

\begin{table}[t]
\small
\centering

\caption{The ablation study on the selected view number in testing on the MultiviewX dataset, training with 3 views and testing with $K$ (2, 3, 4, and 5) views.}
\label{table:viewNumUsingV3CkptMultiviewX}
\begin{tabular}{c@{\hspace{0.0cm}}|c@{\hspace{0.10cm}}c@{\hspace{0.12cm}}c@{\hspace{0.12cm}}c@{\hspace{0.12cm}}c@{\hspace{0.12cm}}|c@{\hspace{0.12cm}}c@{\hspace{0.12cm}}c@{\hspace{0.12cm}}c@{\hspace{0.12cm}}c@{\hspace{0.12cm}}}
\hline
    Method     & \multicolumn{5}{c|}{AVS\_$S_{density}$} & \multicolumn{5}{c}{MVSelect} \\
    View        & MA.    & MP.       & P      & R  & F1    & MA.    & MP.       & P      & R    & F1  \\
\hline
   2 &82.8 &80.4 &97.0 &85.5 &90.9 &77.1 &83.4 &96.5 &80.0&87.5\\
   3 &89.2 &82.1 &98.0 &91.0 &94.4 &88.1 &\textbf{89.8} &98.2 &89.7&93.8\\
   4 &92.7 &\textbf{82.6}&98.5 &94.2 &96.3 &91.2 &87.8 &98.3 &92.9&95.5\\
   5 &\textbf{93.4} &82.4 &\textbf{98.6} &\textbf{94.7} &\textbf{96.6} &\textbf{93.1} &88.9 &\textbf{98.6} &\textbf{94.5}&\textbf{96.5}\\	
\hline
\end{tabular}

\end{table} 

\begin{table}[t]
\centering
\small

\caption{
The ablation study on the same training frame number $F=72$ with different frame selection methods. IDdiff consists of the selected frame ID mean and standard deviation.
}
\label{table:differentFrameSel_wildtrack_multiviewx}
\begin{tabular}{c@{\hspace{0.10cm}}|c@{\hspace{0.20cm}}c@{\hspace{0.2cm}}c@{\hspace{0.2cm}}|c@{\hspace{0.2cm}}c@{\hspace{0.2cm}}c@{\hspace{0.2cm}}}
\hline
    Dataset     & \multicolumn{3}{c|}{Wildtrack} & \multicolumn{3}{c}{MultiviewX} \\

Frame Select     & MA.    & F1   & IDdiff & MA. & F1&  IDdiff\\
\hline
   Random &  	77.5	    &	   87.8  	     &24.7$\pm$ 20.7 & 80.7             & 89.6  &  4.9$\pm$4.1      \\
   Uniform &  77.0		    &	   87.5          &25.0$\pm$ 0.0  & 80.8            & 89.6  &  5.0$\pm$0.0      \\
AVS(Ours)  &  \textbf{79.2}& 	\textbf{88.9}   &18.0 $\pm$52.3 &  \textbf{81.2}   &  \textbf{90.0} &   5.0$\pm$6.0     \\

\hline
\end{tabular}
\vspace{-0.4cm}
\end{table}

\begin{table}[t]
\centering
\small

\caption{
\lbreb{Model cost comparison on MultiviewX.}
}
\label{table:training_cost}
\begin{tabular}{
l@{
\hspace{0.1cm}}|c@{\hspace{0.1cm}}c@{\hspace{0.1cm}}c@{\hspace{0.1cm}}c@{\hspace{0.1cm}}
}
\hline
    Method                  & Memory(GB) & FLOPs(G)  & Train(s) & Test(s)\\
\hline
    MVSelect                    & \textbf{16.879}   & 532.200   & \textbf{1200}  & 10 \\
    Random                      & 17.594   & \textbf{530.703}   & 1700  & \textbf{7} \\
    Random (Pseudo)             & 17.594   & \textbf{530.703}   & 7600  & \textbf{7} \\
    IVS\_$S_g$ (Ours)           & 17.594   & \textbf{530.703}   & 8006  & \textbf{7}  \\
    AVS\_$S_{mask}$ (Ours)      & 17.594   & \textbf{530.703}   & 9172  & \textbf{7}  \\
    AVS\_$S_{density}$ (Ours)   & 17.594   & \textbf{530.703}   & 9194  & \textbf{7}  \\
\hline
\end{tabular}
\vspace{-0.6cm}
\end{table}
\section{Conclusion}
\label{sec:conclusion}

In this paper, we focus on the view selection issue for scene-level multi-view crowd counting and localization tasks. We first propose the independent view selection baseline (IVS), which considers view and scene geometries. Then, based on IVS, we propose the active view selection method (AVS), which incorporates downstream model predictions into view selection and jointly 
optimizes
both view selection and downstream tasks. Extensive experiments on the two tasks demonstrate the advantages of the proposed AVS method over all comparison methods. The proposed method can be applied to novel scenes with limited labels, demonstrating its superior generalization and broader applicability. In the future, the method could also be extended to other BEV-based or 3D reconstruction tasks to reduce labeling costs.




\section*{Acknowledgements}
This work was supported in parts by NSFC (62202312), City University of Hong Kong Strategic Research (7005665), Shenzhen Science and Technology Program (KJZD20240903100022028), and Scientific Development Funds from Shenzhen University.

\bibliographystyle{IEEEtran}
\bibliography{main}

\begin{IEEEbiographynophoto}{\textbf{Qi Zhang}}
received 
the Ph.D. degree in Computer Science from the City University of Hong Kong, Hong Kong SAR, China, in 2021. He is currently a Tenured Associate Professor at the Visual Computing Research Center, College of Computer Science and Software Engineering, Shenzhen University, Shenzhen, China. His research interests include crowd analysis and simulation, 3D reconstruction and generation, autonomous driving, and urban scene point cloud analysis.
\end{IEEEbiographynophoto}
\vspace{-0.5cm}

 \vspace{-0.5cm}

\begin{IEEEbiographynophoto}{\textbf{Bin Li}}
received the B.E. degree in Software Engineering from Jiangxi University of Science and Technology in 2022, China. He is currently pursuing the M.S. degree at the Visual Computing Research Center, College of Computer Science and Software Engineering, Shenzhen University. His research interests include 3D vision and crowd analysis.
\end{IEEEbiographynophoto}
\vspace{-0.5cm}

\begin{IEEEbiographynophoto}{\textbf{Antoni B. Chan}} is currently a Full Professor in the Department of Computer Science, City University of Hong Kong, and is also the Associate Dean of the College of Computing, City University of Hong Kong, Hong Kong SAR, China. His research interests include computer vision, machine learning, pattern recognition, and music analysis.
\end{IEEEbiographynophoto}
\vspace{-0.5cm}

\begin{IEEEbiographynophoto}{\textbf{Hui Huang}}
is a Distinguished TFA Professor of Shenzhen University, where she directs the Visual Computing Research Center and is also the
Dean of the College of Computer Science and Software Engineering, Shenzhen University, China. Her research interests span computer graphics, 3D vision, and visualization. 
\end{IEEEbiographynophoto}

\newpage

\end{document}